\makeatletter
\@namedef{ver@flushend.sty}{9999/12/31}
\makeatother
\documentclass[letterpaper,twocolumn,10pt]{article}
\usepackage{usenix2019_v3}

\usepackage[ruled,vlined]{algorithm2e}

\usepackage{tikz}
\usepackage{amsmath}
\usepackage{amsfonts}
\usepackage{algorithmic}

\usepackage{multirow}
\usepackage[normalem]{ulem}
\usepackage{filecontents}

\usepackage{subcaption}
\usepackage[normalem]{ulem}
\usepackage{graphicx}
\usepackage{array}
\usepackage{soul}
\usepackage{balance}

\newcolumntype{L}[1]{>{\raggedright\let\newline\\\arraybackslash\hspace{0pt}}m{#1}}
\newcolumntype{C}[1]{>{\centering\let\newline\\\arraybackslash\hspace{0pt}}m{#1}}
\newcolumntype{R}[1]{>{\raggedleft\let\newline\\\arraybackslash\hspace{0pt}}m{#1}}

\newcommand{\bz}{B\a'ezier}

\DeclareMathOperator*{\argmax}{arg\,max}

\begin{document}
%-------------------------------------------------------------------------------

%don't want date printed
\date{}

% make title bold and 14 pt font (Latex default is non-bold, 16 pt)
% \title{\Large \bf GAS: Genetic Adversarial Scratches for Neural Networks}
\title{\Large \bf Scratch that! An Evolution-based Adversarial Attack against Neural Networks}

%for single author (just remove % characters)
% \author{
%     {\rm Anonymous Submission}
% }

\author{
    {\rm Malhar Jere}\\
    University of California San Diego\\
    \and
    {\rm Loris Rossi}\thanks{Work done when the author was a Visiting Graduate Student at University of California San Diego}\\
    Politecnico di Milano\\
    \and
    {\rm Briland Hitaj}\\
    SRI International\\
    \and
    {\rm Gabriela Ciocarlie}\\
    SRI International\\
    \and
    {\rm Giacomo Boracchi}\\
    Politecnico di Milano\\
    \and
    {\rm Farinaz Koushanfar}\\
    University of California San Diego\\
}
% \author{
% {\rm Your N.\ Here}\\
% Your Institution
% \and
% {\rm Second Name}\\
% Second Institution
% % copy the following lines to add more authors
% % \and
% % {\rm Name}\\
% %Name Institution
% } % end author

\maketitle

%-------------------------------------------------------------------------------
\begin{abstract}
%-------------------------------------------------------------------------------
We study black-box adversarial attacks for image classifiers in a constrained threat model, where adversaries can only modify a small fraction of pixels in the form of scratches on an image. We show that it is possible for adversaries to generate localized \textit{adversarial scratches} that cover less than $5\%$ of the pixels in an image and achieve targeted success rates of $98.77\%$ and $97.20\%$ on ImageNet and CIFAR-10 trained ResNet-50 models, respectively. We demonstrate that our scratches are effective under diverse shapes, such as straight lines or parabolic B\a'ezier curves, with single or multiple colors. In an extreme condition, in which our scratches are a single color, we obtain a targeted attack success rate of $66\%$ on CIFAR-10 with an order of magnitude fewer queries than comparable attacks. We successfully launch our attack against Microsoft's Cognitive Services Image Captioning API and propose various mitigation strategies.
\end{abstract}

%-------------------------------------------------------------------------------

\section{Introduction}

Deep Neural Networks (DNNs) have achieved state-of-the-art results in image classification~\cite{krizhevsky2012imagenet, simonyan2014very, he2016deep, szegedy2016rethinking, huang2017densely}, machine translation~\cite{devlin2018bert, edunov2018understanding, radford2019language}, and reinforcement learning tasks~\cite{silver2016mastering, bansal2017emergent, vinyals2019grandmaster, silver2017mastering}, and have been adopted for a wide array of tasks, such as cancer detection~\cite{shen2015multi, albarqouni2016aggnet}, malware detection~\cite{vinayakumar2019robust, le2018deep, raff2018malware}, speech recognition~\cite{hannun2014deep}, and more. However, despite these accomplishments, DNNs are surprisingly susceptible to deception by adversarial samples~\cite{szegedy2013intriguing}, which consist of images containing sub-perceptual noise, invisible to humans, that can lead a neural network to misclassify a given input. The adversarial samples can be: (1) targeted (i.e., the goal is to make the model classify the samples to an adversarially desired outcome), or (2) untargeted (i.e., the misclassification can be attributed to any class that is not the ground truth). Researchers have demonstrated that adversarial samples that transfer to the real-world can be printed on paper or 3D-models~\cite{athalye2017synthesizing}, and can target classifiers in different domains, such as images~\cite{szegedy2013intriguing}, speech~\cite{carlini2018audio}, and text~\cite{alzantot2018generating}. 
% Beyond the security threats that adversarial samples pose to neural networks, they offer a fascinating insight into their strengths, weaknesses, and failure modes.
%
\begin{figure}[!t]
    \begin{subfigure}{.5\linewidth}
      \centering
      \includegraphics[width=0.95\linewidth]{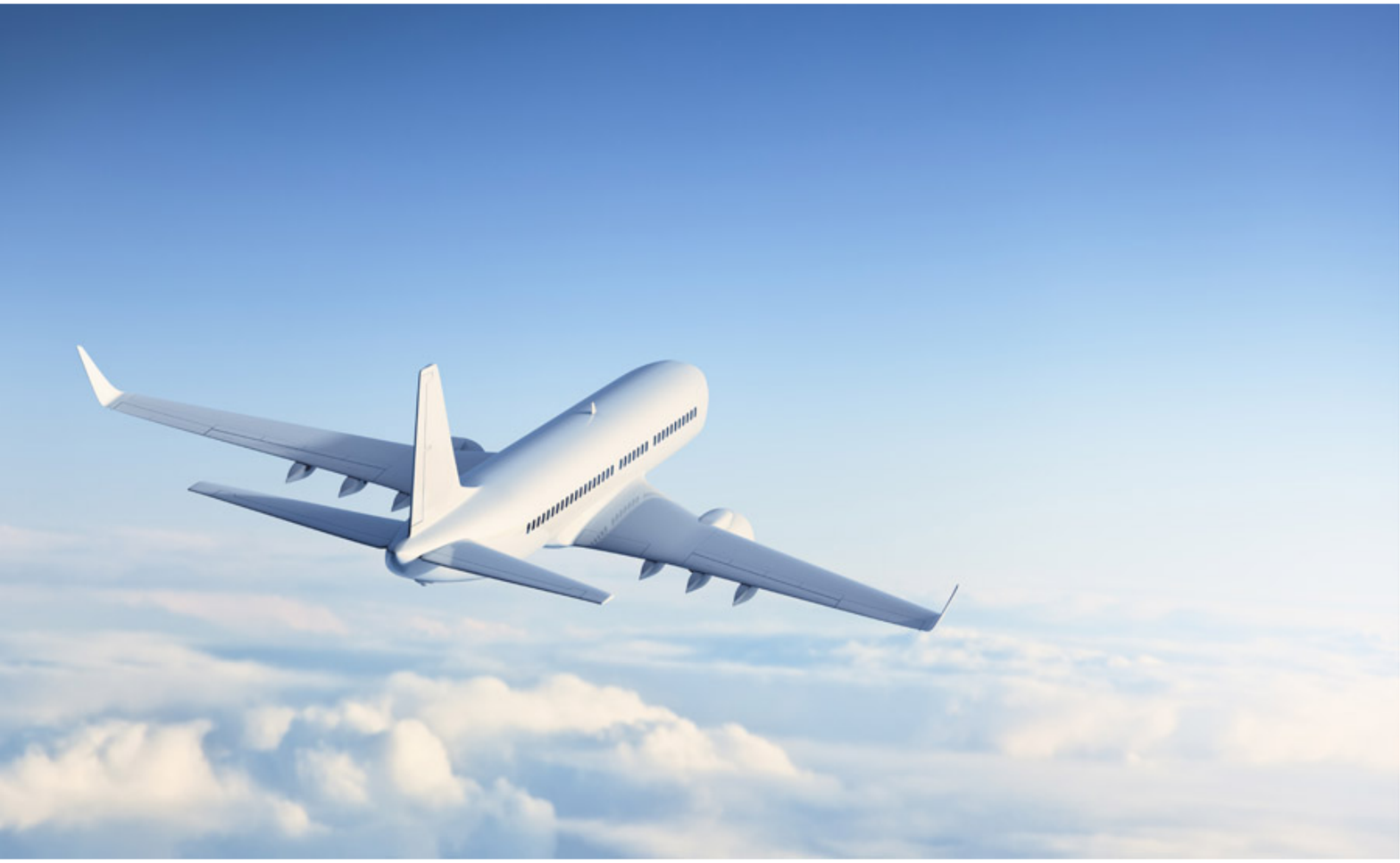}
      \caption{ Original Image.}
      \label{fig:sfig1}
    \end{subfigure}%
    \begin{subfigure}{.5\linewidth}
      \centering
      \includegraphics[width=0.95\linewidth]{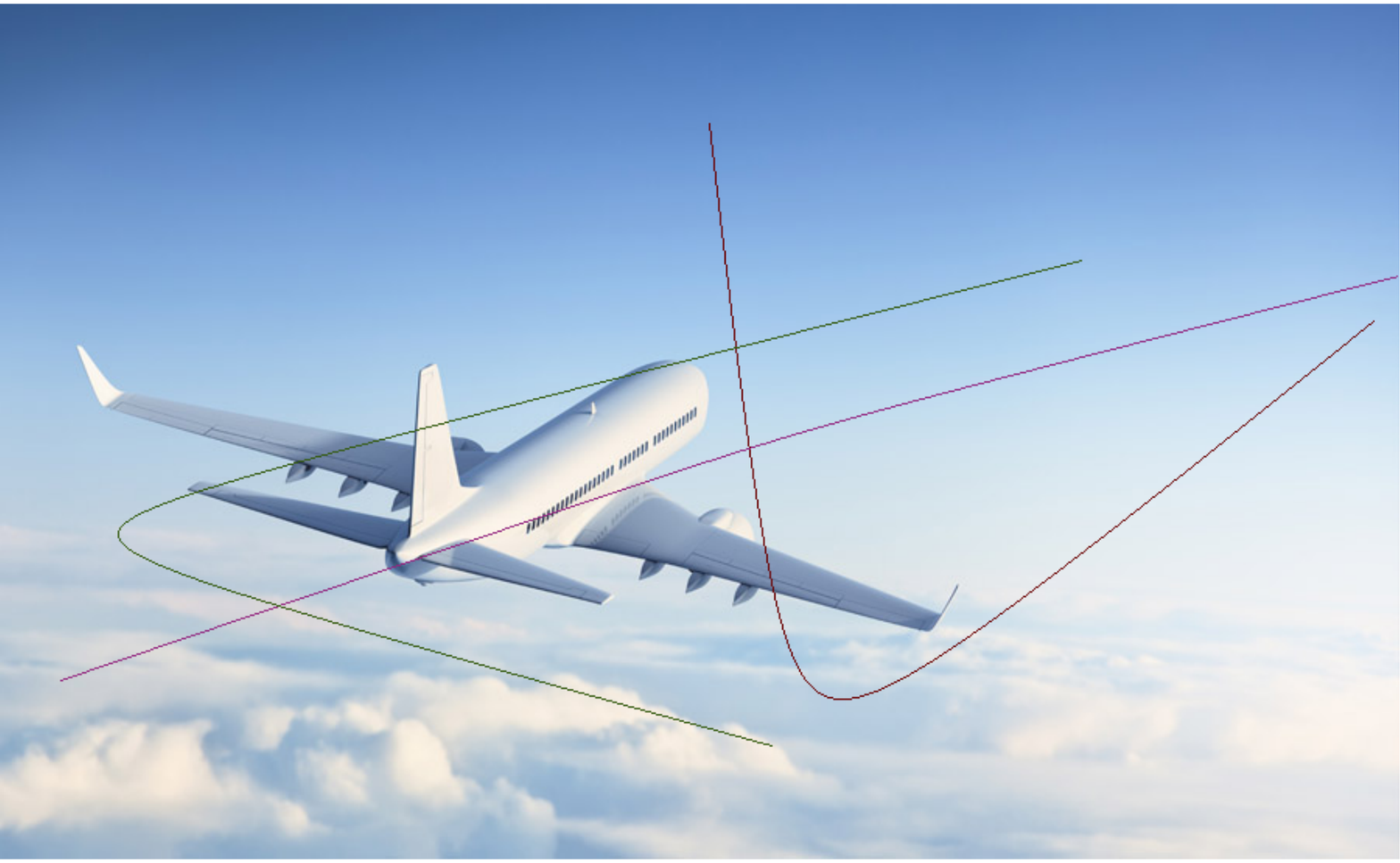}
      \caption{Adversarial Image}
      \label{fig:sfig2}
    \end{subfigure}
    \caption[]{Adversarial attack on Microsoft's Cognitive Services Image Captioning API. Generated caption for Image~\ref{fig:sfig1}: `\emph{a large passenger jet flying through a cloudy blue sky}'. Generated caption for Image~\ref{fig:sfig2}: `\emph{a group of people flying kites in the sky}'. Best viewed in color.\protect\footnotemark}
    \label{fig:vision-attack}
\end{figure}
\footnotetext{An anonymized video demonstration of the attack can be found here:~\url{https://www.youtube.com/watch?v=WminC14RLeY}}

\noindent The majority of existing adversarial sample-generation techniques for image classifiers focuses on invisible perturbations that often cover the entire image, where the maximum perturbation added to the image is upper bounded according to an $L_{p}$-norm metric.
% \gi{it is not very clear to me what $L_{p}$-norm around the image means. Do you mean there is an upperbound on the $L_{p}$-norm of the difference between $x$ and $x'$?}. 
These include the $L_{0}$ norm~\cite{papernot2016limitations}, which captures the number of perturbed image pixels; the $L_{1}$~\cite{moosavi2016deepfool,chen2018ead} norm, which measures the Manhattan norm or absolute sum of perturbation values; the $L_{2}$ norm~\cite{carlini2017adversarial}, which measures the Euclidean distance of the perturbation; or the $L_{\infty}$ norm~\cite{madry2018towards, goodfellow2014explaining}, which is the largest perturbed value. 

In most of these threat models, the adversary has white-box access to the model parameters, i.e., the adversary has full knowledge of the model parameters and training data. Consequently, in these cases, the adversary can easily calculate gradients with respect to the inputs. 
% \bh{Needs rephrasing:}
While these attacks exemplify the significant risks to which neural networks might be exposed, they often do not reflect a realistic threat scenario against computer vision models deployed in the real world, which is often a black-box scenario, where adversaries have limited queries and no access to the model architecture or parameters.

% \begin{figure*}[h!]
%     \centering
%     \begin{subfigure}{.5\linewidth}
%       \centering
%       \footnotesize \textsf{Original Image \\}
%       \includegraphics[width=0.3\linewidth]{figures/examples/tiger.jpeg}
%     \end{subfigure}%
%     \begin{subfigure}{.5\linewidth}
%       \centering
%       \footnotesize \textsf{1 straight-line \\ scratch \\}
%       \includegraphics[width=0.3\linewidth]{figures/examples/tiger_1_line_scratch.pdf}
%     \end{subfigure}
%     \begin{subfigure}{.5\linewidth}
%       \centering
%       \footnotesize \textsf{2 straight-line \\ scratches \\}
%       \includegraphics[width=0.3\linewidth]{figures/examples/tiger_2_line_scratch.pdf}
%     \end{subfigure}
%     \\[10pt]
    
%     \begin{subfigure}{.5\linewidth}
%       \centering
%       \footnotesize \textsf{1-bezier curve \\ scratch \\}
%       \includegraphics[width=0.4\linewidth]{figures/examples/tiger_1_bezier_scratch.pdf}
%     \end{subfigure}%
%     \begin{subfigure}{.5\linewidth}
%       \centering
%       \footnotesize \textsf{2-bezier curve \\ scratch \\}
%       \includegraphics[width=0.4\linewidth]{figures/examples/tiger_2_bezier_scratch.pdf}
%     \end{subfigure}
%     \begin{subfigure}{.5\linewidth}
%       \centering
%       \footnotesize \textsf{3-bezier curve \\ scratch \\}
%       \includegraphics[width=0.4\linewidth]{figures/examples/tiger_3_bezier_scratch.pdf}
%     \end{subfigure}
%     \\[10pt]

%     \caption{Variable location, double scratch attacks on ImageNet in the Image domain.}
%     \label{fig:img-attacks}
% \end{figure*}

\begin{figure*}[ht!]
    \centering
    \includegraphics[width=\textwidth]{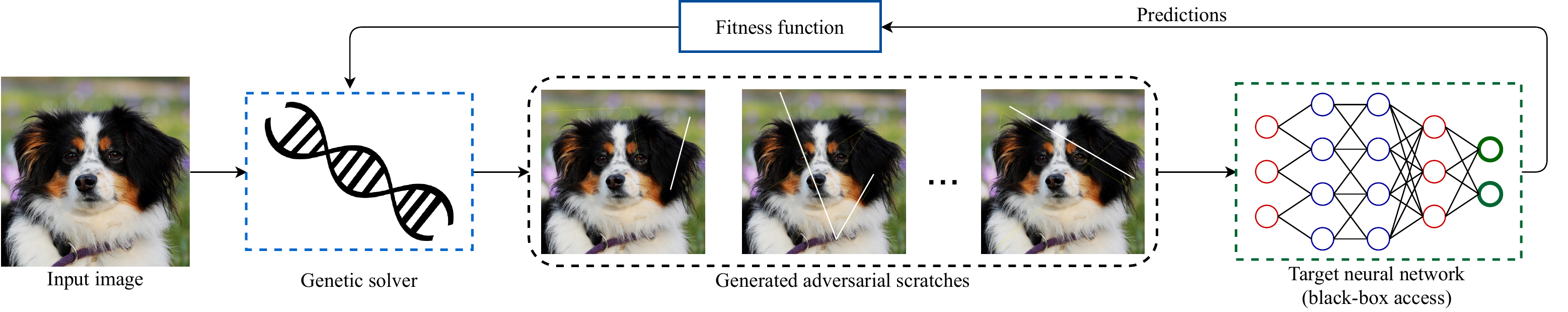}
    \caption{Framework to generate variable location adversarial scratches. Scratch parameters include: (1) number of scratches, (2) domain of scratches (image or network domain), (3) location parameters of scratch (fixed or variable), (4) population size, (5) number of iterations, (6) mutation and crossover rate. Best viewed in color.} 
    % \bh{I noticed that the schema figure is not referenced in the text.}
    \label{fig:scratch_schema}
\end{figure*}

This paper proposes a new parametric $L_{0}$ black-box adversarial attack against machine learning (ML) models for this limited, but realistic threat scenario. In the threat model, the adversary can only modify pixels in a constrained manner. We call these modifications adversarial scratches, inspired by the scratches commonly found in actual printed images.
%in the form of \textit{scratches} for an image. 
Our scratches are parametrized as simple geometric shapes: line segments or second-degree B\a'ezier curves, which are evolved through Evolution Strategies, as highlighted in Figure~\ref{fig:scratch_schema}. We furthermore enforce a query budget to simulate rate limiting for real-world ML systems.
% \gi{is this part of the method or of results comments? I would give more relevance to the key components of our approach such as the fact we are using geometric model for scratches that evolves through GA, which are perfectly fit to the considered black-box settings (as opposed of the method mentioned before which require access to the network parameters}.

We propose two novel evolutionary black-box neural network attacks, for which the only available information is the confidence for an image caption or for individual class labels.
% \bh{\textbf{Consider merging this with point 5 of contributions!} Furthermore, we test our algorithm against Microsoft's commercially available Cognitive Services Image Captioning API and are successfully able to fool it into producing wrong captions or even producing no captions at all, with an example of a successful attack being demonstrated in Figure~\ref{fig:vision-attack}.}
% We furthermore observe that our attacks are robust even under defenses such as JPEG compression.

% Our contributions include:
\subsection{Contributions}
\begin{itemize}
    \item \textit{Adversarial scratches}, a new class of constrained black-box attacks for image classifiers. Adversarial scratches leverage Evolution Strategies, where scratches are evolved parametrically according to a pre-defined fitness function, and can successfully attack ResNet-50, VGG-19, and AlexNet architectures trained on the CIFAR-10 and ImageNet datasets.
    % \bh{consider expanding this a bit with EAs.}
    
    \item \textit{Attack effectiveness evaluation} across a variety of attack parameters, including query dependence, scratch shape, scratch location, and scratch color. 
    Even in the extremely constrained case with a single color for a single scratch, our attacks achieved a $66\%$ targeted success rate for CIFAR-10 trained neural networks.
    
    % \item \bh{We successfully deploy our attacks in benchmark models, e.g., Resnet-50, VGG-16, ... trained on CIFAR-10 or ImageNet datasets.}

    \item \textit{Successful attack deployments} in both image and network domains, similar to LaVAN work by Karmon et al.~\cite{karmon2018lavan}. The image-domain scratches are restricted to lie within the normal dynamic image range of $[0,1]$; the network-domain scratches are not necessarily restricted on predefined bounds.
    %the network-domain scratches are not similarly restricted. 
    
    \item \textit{A real-world attack} against Microsoft's commercially available Cognitive Services Image Captioning API. Our attacks successfully fooled the API into producing wrong captions or no captions (an example of a successful attack is demonstrated in Figure~\ref{fig:vision-attack}), thereby demonstrating a real-world manifestation of an adversarial attack on ML systems. We have contacted Microsoft regarding this vulnerability.
    
    \item \textit{Countermeasures} to our attacks. We propose, evaluate, and discuss different countermeasures to our attacks including,
    JPEG compression, median filtering, and image clipping, and we assess their effectiveness in mitigating the effect of adversarial scratches and their impact on benign images.
    % \highlight{LR: I would make two separate items, one for the API attacks and one for the defenses, so that the attack to Microsoft's API is more evident and we can emphasize that we attacked a real-world system}
    
    % \item Finally, we propose and evaluate a set of defenses such as JPEG compression and median filtering that can improve model robustness to our adversarial scratches.
\end{itemize}

% Our findings motivate the need to explore further defenses against $l_{0}$ attacks.

\subsection{Organization}
The rest of the paper is organized as follows: In Section~\ref{sec:background}, we discuss related work and provide the necessary background information. Section~\ref{sec:newmethodology} discusses the methodology pursued to generate adversarial scratches. Section~\ref{sec:experimental_setup} and Section~\ref{sec:results} depict our experimental setting and our results, respectively. In Section~\ref{sec:discussion}, we further elaborate on our results, initial failed attempts, and potential defense strategies. We provide our conclusions and future directions in Section~\ref{sec:conclusion}.
%enlisting promising future directions.

% \bh{Write this in the end when the sections order is in place.}

% !TEX root = usenix2019_v3.1.tex
\section{Background and Related Work}
\label{sec:background}

There is extensive prior work on generating adversarial samples for neural networks across multiple input domains. We highlight the different types of adversarial attacks on neural networks and provide a quick overview of the evolution strategies used in our work. 

% BH here...
\begin{figure*}[!t]
\centering
    \begin{subfigure}{.3\linewidth}
      \centering
      \includegraphics[width=0.92\linewidth]{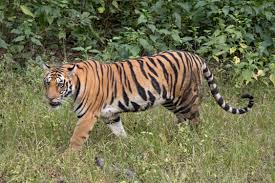}
      \caption{Original Image.}
      \label{fig:tigerfig1}
    \end{subfigure}%
    \begin{subfigure}{.3\linewidth}
      \centering
      \includegraphics[width=0.95\linewidth]{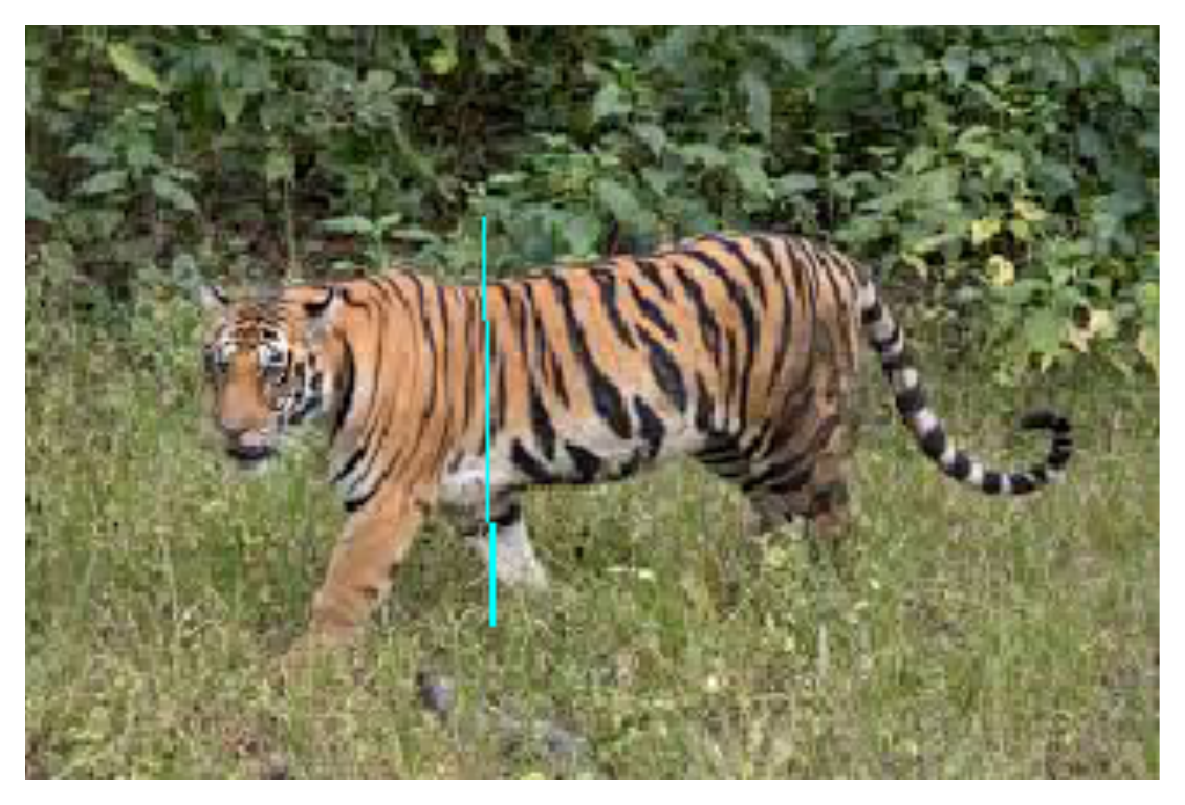}
      \caption{1 straight-line scratch}
      \label{fig:tigerfig2}
    \end{subfigure}%
    \begin{subfigure}{.3\linewidth}
      \centering
      \includegraphics[width=0.95\linewidth]{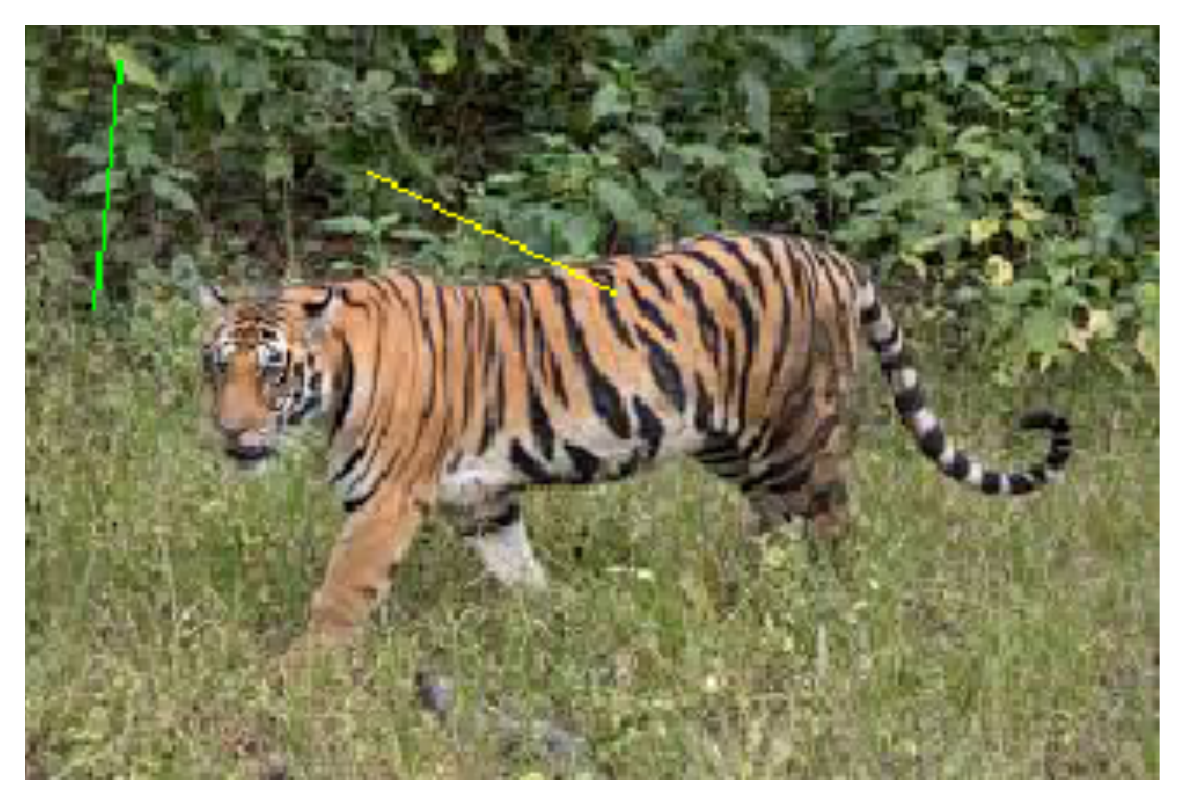}
      \caption{2 straight-line scratches}
      \label{fig:tigerfig3}
    \end{subfigure}
    \\
    \begin{subfigure}{.3\linewidth}
      \centering
      \includegraphics[width=0.95\linewidth]{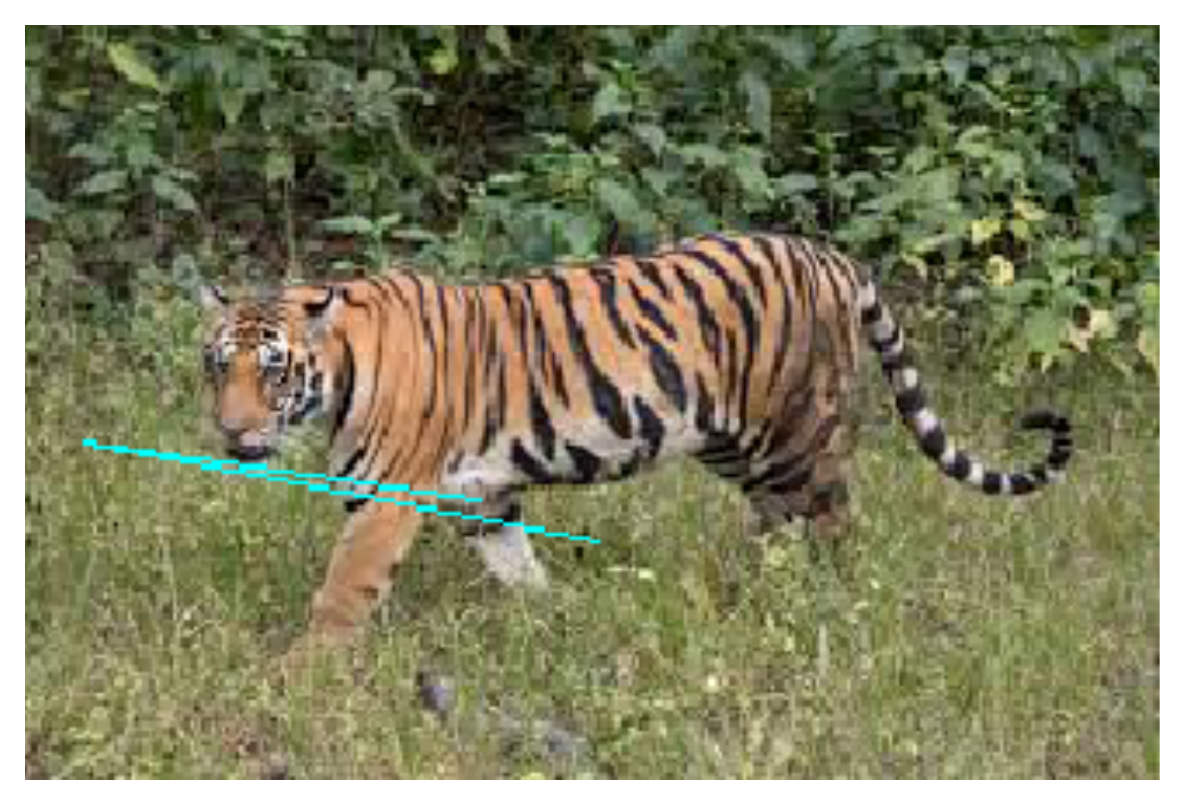}
      \caption{1-B\a'ezier curve scratch}
      \label{fig:tigerfig4}
    \end{subfigure}%
    \begin{subfigure}{.3\linewidth}
      \centering
      \includegraphics[width=0.95\linewidth]{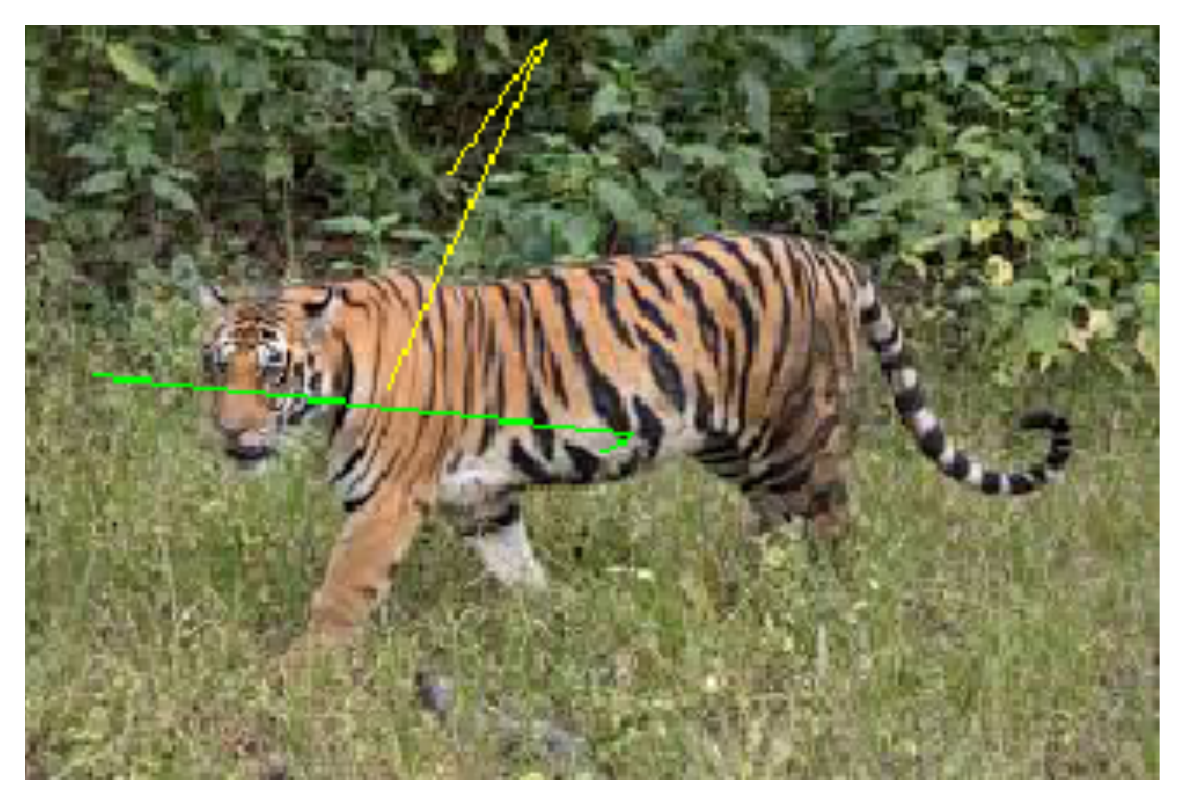}
      \caption{2-B\a'ezier curve scratches}
      \label{fig:tigerfig5}
    \end{subfigure}%
    \begin{subfigure}{.3\linewidth}
      \centering
      \includegraphics[width=0.95\linewidth]{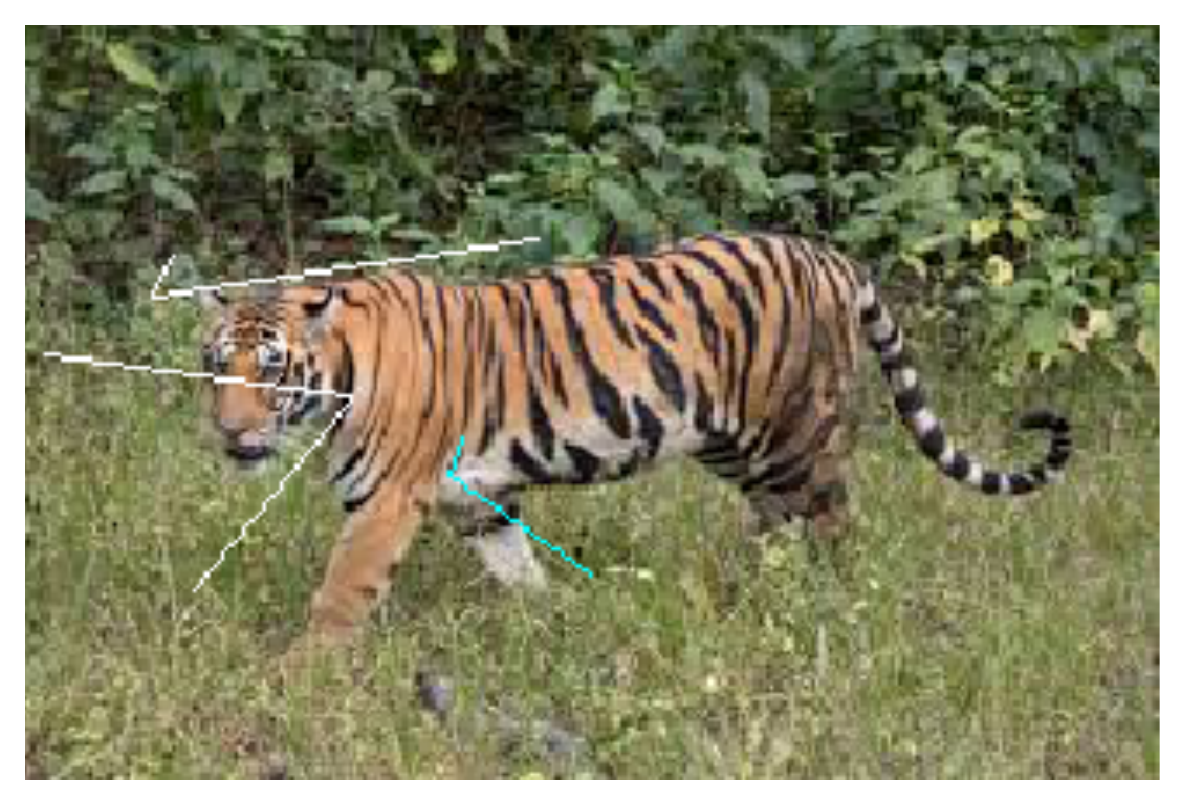}
      \caption{3-B\a'ezier curve scratches}
      \label{fig:tigerfig6}
    \end{subfigure}
    \caption{Adversarial Scratches can have diverse manifestations, such as B\a'ezier Curves or Straight Lines. Individual scratches for the same image may have different colors. Best viewed in color.}
    % Variable location, double scratch attacks on ImageNet in the Image domain. \bh{Please elaborate more on the images displayed. Furthermore, I wonder if there is a way to highlight the scratches. I added a `Best viewed in color' sentence, but even then it is hard. For instance for figures (b) and (d) the scratches are almost invisible.} \highlight{Best viewed in color.}}
    \label{fig:examples-img-attacks}
\end{figure*}

\subsection{Adversarial Attacks on Neural Networks} 

Prior works on adversarial attacks on neural networks consist of methods to deceive~\cite{carlini2017adversarial, papernot2016limitations, athalye2018obfuscated, goodfellow2014explaining} or cause denial-of-service of neural networks~\cite{sponge_attacks}. Our work considers adversarial attacks that attempt to deceive image classifiers.
We consider a neural network $f( \cdot)$ used for classification where $f(x)_{i}$ represents the softmax probability that image $x$ corresponds to class $i$. Images are represented as $x \in
[0,1]^{w.h.c}$, where $w, h, c$ are the width, height, and number of channels of the image. We denote the classification of the network as
$C(x) = \argmax_{i} f(x)_{i}$, 
with $C^{\ast}(x)$ 
representing the ground truth of the image. 
Given an image $x$ and an adversarial sample $x'$, the adversarial sample has the following properties:
\begin{itemize}
    \item $L(x'-x)$ is small for some image distance metric $L$.
    
    \item $C(x') \neq C^{\ast}(x) = C(x)$. This means that the prediction on the adversarial sample is incorrect whereas the original prediction is correct.
\end{itemize}

% \lr{Here we use $C$ as a function, while in Eq. \ref{eq:img_domain_targeted} $C$ is used as a single value. Is it ok?}

% \highlight{Adversarial samples were discovered for Support Vector Machines by Biggio~\cite{biggio2012poisoning}, and were subsequently discovered for Deep Neural Networks were first discovered  by Szegedy et al.~\cite{szegedy2013intriguing}, who then inspired numerous papers that generate adversarial samples for neural networks for both white- and black-box scenarios.}

Biggio et al.~\cite{biggio2012poisoning} were the first to show the occurrence of adversarial examples in Support Vector Machines, while Szegedy et al.~\cite{szegedy2013intriguing} discovered that deep neural networks are prone to such attacks as well. Since then, there has been a plethora of works generating adversarial samples for both white and black-box scenarios.

% Adversarial samples were first discovered by Biggio et al. for Support Vector Machines~\cite{biggio2012poisoning}, following which Szegedy discovered adversarial samples for Deep Neural Networks~\cite{szegedy2013intriguing}. Since then there has been a plethora of works generating adversarial samples for both white and black-box scenarios.

\noindent\textbf{White-Box Adversarial Attacks.}
In the white-box threat scenario, adversaries have full access to the neural network parameters and can calculate gradients with respect to inputs. Most white-box adversarial sample attacks characterize robustness with respect to the $L_{p}$ norm, as outlined below.
% \gi{is this part about norms redundant w.r.t the previous section?}

\begin{itemize}
    \item \textbf{$L_{0}$-norm} attacks limit the number of perturbed pixels. Several examples include the JSMA attack~\cite{papernot2016limitations}, which finds vulnerable pixels through saliency maps for inputs, SparseFool~\cite{modas2019sparsefool}, which exploits the low mean curvature of decision boundaries, and~\cite{narodytska2017simple}, which uses greedy search to find adversarial perturbations.
    
    \item \textbf{$L_{1}$-norm} attacks place an upper bound on the Manhattan distance or absolute sum of perturbation values. Examples include DeepFool~\cite{moosavi2016deepfool} and the EAD attack~\cite{sharma2017attacking}.
    
    \item \textbf{$L_{2}$-norm} attacks place an upper bound on the Euclidean distance of the adversarial perturbation. A notable example is the Carlini-Wagner L2~\cite{carlini2017towards} attack, which performs gradient descent on the predicted perturbation norm.
    
    \item \textbf{$L_{\infty}$-norm} attacks place an upper bound on the maximum perturbation that can be applied to each pixel. Notable examples include the Fast Gradient Sign Method~\cite{goodfellow2014explaining} and the Projected Gradient Descent method~\cite{madry2018towards}.
\end{itemize}

In addition to $L_{p}$-norm constrained attacks, several other attacks use gradient information, such as LaVAN~\cite{karmon2018lavan} and adversarial patches~\cite{brown2017adversarial}.

\noindent\textbf{Black-box Adversarial Attacks.}
In the black-box threat scenarios adversaries possess limited information of the parameters of the neural network under attack. This threat model may range from adversaries having access to the full prediction vector, i.e., \textit{soft-label}, to simply having access to the top predicted label, i.e., \textit{hard-label}. Black-box attacks are generally more challenging to implement than their white-box counterparts due to limited information available to the adversary.

Recent literature has proposed numerous black-box attacks that restrict perturbations with respect to an $L_{2}$ or $L_{\infty}$ norm constraint. Narodytska and Kasiviswanathan~\cite{narodytska2017simple} use a naive policy of perturbing random segments of an image to generate adversarial examples. Bhagoji et al.~\cite{bhagoji2017exploring} reduce the feature space dimensions using PCA and random feature grouping before estimating gradients. Chen et al.~\cite{chen2017zoo} use finite differences to estimate gradients for a gradient based-attack. Ilyas et al.~\cite{ilyas2018black} use Natural Evolutionary Strategies to estimate gradients, while GenAttack~\cite{alzantot2018genattack} uses genetic algorithms to estimate gradients. Guo et al.~\cite{guo2018low} identify that low-frequency perturbations allow for improved query efficiency for black-box attacks, which is further evaluated in SimBA~\cite{guo2019simple}. Lin et al. utilize differential evolution to generate $L_{2}$-norm bounded adversarial perturbations~\cite{lin2020black}.

In contrast to these works, $L_{0}$-norm bounded black-box attacks are rarely explored. The One-Pixel Attack~\cite{su2019one} uses Differential Evolution to find the optimal $(x,y)$ pixel coordinates and RGB values, but it has a poor success rate on larger neural networks and requires several thousand queries. The PatchAttack~\cite{yang2020patchattack} method uses reinforcement learning to  optimally place pre-generated textured patches, however, the patches are often very large and highly visible, covering up to $20\%$ of the entire image. %\bh{Update this based on Thursday's discussion. Potentially some of these points can go to contributions: In this vein, our work is most similar to that of Jere et al.~\cite{jere2019scratch} who use CMA-ES to find perturbations in the form of fixed scratches along an image; however, we make several key distinctions in this work.}
% Mention that we also use differential evolution, and implement a physical attack also. Also used a parametric model for the perturbation which vastly reduces number of search parameters, as compared to previous methods which try to simply reduce the raw perturbation. - add this in primary contributions. 
The Adversarial scratches discussed in our work use Differential Evolution, similar to the One-Pixel Attack, but proposes a parametric model for the perturbation which vastly reduces the number of search parameters. Finally, our approach implements a physical attack on a real-world image classification framework. 

%%%%%%%%%%%%%%%%%%THIS WE CAN ADD IN THE DISCUSSION ABOUT PREVIOUS SUBMISSIONS
% \begin{itemize}
%     \item We evaluate our scratches for both Image and Network Domains, while their scratches are strictly in the Network Domain.
%     % LR: here below I wrote down some suggestions
%     \item We evaluate the robustness of models under a number of scratches, while their evaluations only consist of 1 scratch.
%     % Here we use from 1 to 4 scratches per attack, while on their scratches only 1 scratch is used
%     \item We evaluate a diversity of scratch manifestations, namely straight lines and second-order Bezier curves, while they only evaluated their attack for straight lines.
%     % Here we evaluated straight lines and second-order Bezier curves, while they used only straight lines
%     \item We extend our attack to the CIFAR-10 dataset as well as a physical attack on a real-world Machine Learning System, rather than evaluating solely on ImageNet.
%     % Here we also considered CIFAR-10, other than ImageNet
%     \item We investigated the effect of numerous scratch parameters, such as single color, multiple colors, scratch location and scratch shape as opposed to their work where each scratch pixel was a different value.
%     % Here we evolved also the location of the scratch, and in some cases we used only a single color per scratch, as opposed to their work where every scratch's pixel has a different value
% \end{itemize}

\subsection{Evolution Strategies}
% \gi{is it better to introduce in the meantime the notation used in algorithm 1? Please check there is a notation clash with $f$ which is both the network and the fitness function of GA}
% \highlight{LR: is it "evolutionary" or "evolution" strategy? I've only seen "evolution" strategy, like in \cite{hansen2016cma} or \url{https://openai.com/blog/evolution-strategies/}. Let's make a choice in order to be consistent throughout the paper}
The generation of adversarial scratches uses evolution strategies, which are population-based gradient-free optimization strategies. They are roughly inspired by the process of natural selection, in which a population \textit{P} of candidate solutions is generated at each iteration, termed a \textit{generation}. Candidate solutions are evaluated using a \textit{fitness function}, and "fitter" solutions are more likely to be selected to breed the next generation of solutions.
The whole process terminates either when a candidate solution satisfies some pre-defined criteria, or when the maximum number of generations is reached.
% \lr{I would add: The whole process terminates either when a candidate solution satisfy some pre-defined criteria, or when the maximum number of generations is reached.} 
The next generation is generated through a combination of \textit{crossover} and \textit{mutation}. Crossover involves taking numerous parent solutions, combining their parameters, and generating a new generation of solutions, similar to biological reproduction. Mutation applies small random perturbations to population members to increase the diversity of population members and provide a better exploration of the search space. Evolution strategies are particularly suited for black-box optimization, as they do not require any information about the underlying fitness function.
% \highlight{LR: Maybe we can also say that evolution strategies are particularly suited for black-box optimization, since they need nothing but a fitness function}
% Algorithm ~\ref{genetic} outlines a general framework to solve optimization problems with evolutionary strategies.
% \begin{algorithm}
% \caption{Framework for solving optimization problems with evolutionary strategies}
% \label{genetic}
% \begin{algorithmic}
% \REQUIRE $N$, $solver$, $REQUIRED\_FITNESS$
% \WHILE{$True$}
% \STATE $solutions$ $=$ $solver.ask()$
% \STATE $fitness\_list = zeros(N)$
% \FOR{$i=0$ to $N$}
% \STATE $fitness\_list[i]$ = $fitness\_function(solutions[i])$
% \ENDFOR
% \STATE $solver.tell(fitness\_list)$
% \STATE $best\_solution, best\_fitness = solver.result()$
% \IF {$best\_fitness > REQUIRED\_FITNESS$}
% \RETURN $best\_solution$
% \ENDIF
% \ENDWHILE
% \end{algorithmic}
% \end{algorithm}
Our work uses the Differential Evolution (DE) and Covariance-Matrix Adaptation Evolution Strategies (CMA-ES) algorithms.

\noindent\textbf{Differential Evolution.}
% Differential Evolution is an evolution strategy that allows for constrained optimization in that algorithm parameters are constrained to be within a certain bound. 
Differential Evolution is an evolution strategy that allows for constrained optimization of algorithm parameters within a certain bound. This is highly advantageous for generating adversarial scratches for real-world scenarios where perturbations are restricted to be in the dynamic range of images. Algorithm~\ref{algorithm:differential_evolution} highlights the steps to generate potential candidates for a given set of parameters to the Differential Evolution solver.

% Differential Evolution consists of 3 main stages: mutation, crossover and evaluation. The algorithm first generates a set of $P$ candidates randomly. Then, for each iteration $i$, and for each candidate $x_j$, 3 candidates $a,b,c$ are chosen from $x_{1:P}$ that are unique and distinct from $x_j$. A random index $R$ is chosen from the dimensionality $n$ of $x_j$, and the agent's position $y$ is then generated as follows:
% \begin{itemize}
%     \item For each index $k \in {1,2,..n}$, a uniformly distributed number $r_k$ is chosen from $U(0,1)$. If $r_{k} \leq CR$, then $y_k = a_k + F\times(b_k - c_k)$, else $y_k = x_k$ 
%     \item If $f(y) \leq f(x)$ then we replace the candidate $x_j$ with the improved candidate $y$.
% \end{itemize}

\begin{algorithm}[t]
\caption{Differential Evolution for a fitness function $F$, which draws the adversarial scratch and evaluates the model performance on the scratched image. The algorithm parameters are population size $\lambda$, number of iterations $N$, mutation rate $m$, cross-over rate $CR$, and bounds for each candidate solution $BOUNDS$. It returns candidate $x \in \mathbb{R}^{n}$, where $n$ is the size of the solution vector being evaluated.}
\label{algorithm:differential_evolution}
\begin{algorithmic}
\REQUIRE $F$, $\lambda$, $N$, $m$, $CR$, $BOUNDS$

\STATE $X \leftarrow uniform\_random\_generator(\lambda, n, [0,1])$

\FOR{$i=1$ to $N$}

\FOR {$j=1$ to $\lambda$}

\STATE $x = X_{j}$ \\
\STATE $p,q,r \leftarrow random\_sample(3,n) s.t p \neq q \neq r$
\STATE $a,b,c \leftarrow X[p, q, r] $ 
% \lr{where is the definition of p, q, r? I guess they are random values from 1 to $\lambda$} \\

\STATE $MUTANT = clip(a + m \times (b-c), BOUNDS)$ \\

\STATE $r = uniform\_random\_generator(n, [0,1])$ \\

\FOR{$k=1$ to $n$ 
% \lr{I think it should be for  $k=1$ to $n$, not $N$}
}

\IF{$r_k \leq CR$}
\STATE{$y_k = MUTANT[k]$}
\ELSE
\STATE{$y_k = x[k]$}
\ENDIF

\ENDFOR

\IF{$F(y) \leq F(x)$}
\STATE{$X_j = y$}
\ENDIF
\ENDFOR

\ENDFOR

\RETURN $x \in X$ such that $F(x) < F(x_k) \ \forall \ k \in \{1,2...N\}$
\end{algorithmic}
\end{algorithm}

% Differential Evolution (DE) is an iterative, population-based optimization algorithm commonly used to solve multidimensional constrained optimization problems with very few underlying assumptions. At each generation, DE generates candidate solutions and attempts to find the candidates that perform best against a given fitness function. Subsequent generations are then derived from the best performing candidates, and the process is repeated till convergence or maximum number of iterations is completed. 

% DE does not utilize gradient information and makes very few  assumptions about the underlying problem, thereby making it an excellent fit for the constrained optimization problem of generating adversarial samples.
\noindent\textbf{CMA-ES.} Covariance Matrix Adaptation Evolution Strategy (CMA-ES) is an evolution strategy used for the optimization of real non-convex functions in a continuous domain and it is considered a state-of-the-art algorithm in evolutionary computation~\cite{hansen2016cma}. At each generation, CMA-ES samples a population of candidate solutions from a multivariate normal distribution, whose mean, standard deviation, and covariance matrix are updated after each iteration. The covariance matrix is used to track pair-wise dependencies of the optimized parameters; this allows the solver to dynamically change the degree of exploration during the optimization process.
%\highlight{MJ: need to add CMA-ES algorithm here in detail. Loris, could you please add it?}
Algorithm~\ref{algorithm:cmaes} shows how CMA-ES works on a high level, with covariance matrix $C$, and evolution paths $p_c$ and $p_\sigma$. For further details on its implementation, please refer to the CMA-ES author's tutorial~\cite{hansen2016cma}.

\begin{algorithm}[ht]
\caption{Covariance Matrix Adaptation Evolution Strategy for a fitness function $F$. The algorithm parameters are population size $\lambda$, number of iterations $N$, mean vector $\mu \in \mathbb{R}^{n}$, and step size $\sigma_{k}$, where $n$ is the size of the solution vector being evaluated. It returns candidate $x_{1} \in \mathbb{R}^{n}$ or $\mu$}
\label{algorithm:cmaes}
\begin{algorithmic}
\REQUIRE $F$, $\lambda$, $N$, $\mu$, $\sigma_{k}$
% \STATE $ $ \\

\STATE \COMMENT{Initialize variables}
\STATE $C = I, p_{\sigma} = 0, p_{c} = 0$
\FOR{$i = 1$ to $N$}
    
    \STATE \COMMENT{Generate new population}
    \FOR{$j = 1$ to $\lambda$}
        \STATE $x_{j} = sample\_from\_N(\mu,\ \sigma^{2}C)$
        \STATE $f_{j} = F(x_{j})$
    \ENDFOR
    
    \COMMENT{Sort best candidates and update all parameters}
    \STATE $x_{1...\lambda} = x_{s(1)...s(\lambda)} \ with \ s(i) = argsort (f_{1...\lambda}, j)$
    \STATE $\mu' = \mu$
    \STATE $\mu \leftarrow update\_means(x_{1},...,x_{n})$
    \STATE $p_\sigma = update\_ps(p_{sigma}, \sigma^{-1}C^{-1/2}(\mu-\mu')$
    \STATE $p_{c} = update\_pc(p_{c}, \sigma^{-1}(\mu-\mu'),\||p_\sigma\||)$
    \STATE $C = update\_c(C, p_{c}, (x_{1} - \mu')/\sigma, ..., (x_{\lambda} - \mu')/\sigma)$
    \STATE $\sigma = update\_sigma(\sigma, ||p_\sigma||)$
\ENDFOR
\RETURN $x_{1}$ or $\mu$
\end{algorithmic}
\end{algorithm}

% !TEX root = usenix2019_v3.1.tex
\section{Methodology}
\label{sec:newmethodology}

In this section we frame the generation of adversarial scratches as a constrained-optimization problem. We highlight the parameters used to generate scratches as well as characterize scratch generation techniques depending on the dynamic range of images.

% \begin{figure*}[ht!]
%     \centering
%     \includegraphics[width=\textwidth]{figures/schema_scratch_v2.pdf}
%     \caption{Framework to generate variable location adversarial scratches. Scratch parameters here include: (1) number of scratches, (2) domain of scratches (image or network domain), (3) location parameters of scratch (fixed or variable), (4) population size, (5) number of iterations, (6) mutation and crossover rate.}
%     \label{fig:my_label}
% \end{figure*}

\subsection{Problem Description}
% Generating adversarial samples can be formalized as a constrained optimization problem.
Formally, we consider an image classifier $f$ represented as $f: [0,1]^{w.h.c} \rightarrow \mathbb{R}^{K}$ where $K$ is  defined as the number of classes. $f(x)_{i}$ represents the softmax probability that input $x$ corresponds to class $i$.  Images are represented as $x \in [0,1]^{w.h.c}$, where $w, h, c$ are the width, height, and number of channels of the image. The goal of the adversary is to produce an adversarial sample $x' \in [0,1]^{w.h.c}$  that satisfies the following properties:
\begin{align*} 
\mathrm{maximize} \ f(x')_{adversarial} \\ 
\mathrm{subject \ to} \ L(x' - x) \leq d
\end{align*}
Here $L(\cdot)$ represents some distance metric for the adversarial perturbation, which is often minimized or upper-bounded. Prior works often involved utilizing the $L_{p}$ norm as the distance metric. In this vein, our work is most similar to $L_{0}$-norm bounded attacks.

% \gi{what follows is no more part of the problem description sub-section, but rather of the "proposed solution". Can we add a subsubsection titled "Proposed Attacks" or a better term? It is important to make clear what is our contribution and take it apart form the problem definition}

% While prior works focus on generating perturbations that cover an entire region of the image, we depart from this standard methodology.
% \gi{is it true that others affect the entire image? Attacks bounding L0 norm shouldn't take the entire region. Maybe it's better to directly go to our contribution: We propose the first attack that is entirely based on scratches. In particular...} 

\subsection{Proposed Solution}
We propose the first parametric black-box $L_{0}$ attack where the adversarial perturbations are added in the form of scratches on an image. In particular a perturbation $\delta \in \mathbb{R}^{w.h.c}$ is confined over a small area of the image $x$ defined by the support of the scratch, the scratch mask $m \in \{0,1\}^{w.h}$. Scratches are defined as 
\begin{equation}
x' = (1- m) \odot  x + m \odot \delta,
\end{equation}
where $\odot$ denotes the element-wise multiplication.
% , and $\delta \in [0,1]^c$ is the scratch color. %a mask $m \in \{0,1\}^{w.h}$ such that $x' = (1- m) \odot  x + m \odot \delta$, where $\odot$ is element-wise multiplication.

Scratches can be manifested either as B\a'ezier curves~\cite{farin2002handbook} or as line segments, and can consist of a single color for all pixels belonging to a scratch or have distinct colors for each pixel. Ours is the first attack to adopt a parametric model to define the scratch mask $m$, which is key to reducing the number of attack parameters. Adversarial scratches often look like natural artifacts but can significantly affect the prediction of neural networks. Parametric modeling of the scratches allows for flexible attacks on an image.

% \gi{Adopt a parametric model to define the scratch mask $m$, is key to reduce the number of parameters our attacking procedure has to optimize. Parametric models also result in scratches featuring regular shapes, that on the one hand look like a natural artifact, on the other hand introduce patterns that can strongly affect the network output. Moreover, our modeling choice results in a very flexible attack, since we can generate $m$ via Differential Evolution or keep it fixed, depending on the attacking scenario.}

Depending on the attack scenario, we choose to keep the mask $m$ fixed or to evolve its parameters via Differential Evolution. 
% Prior works on generating adversarial samples typically characterize utilize the $L_{p}$ norm as this metric, where $p=0,1,2,\infty$. Our work is most similar to $L_{0}$ attacks, where we limit the number of pixels that can be modified. We set the perturbation $\delta$ to be confined over a small area of the image $x$, and replace this area rather than add perturbations to it. We achieve this by obtaining a mask for pixels $m \in \{0,1\}^{w.c}$ according to a particular scratch equation and replacing the pixels in $x[m]$ with $\delta$. 
% Furthermore, we explore the behavior of \textit{multiple} adversarial scratches and their effects on model robustness. 
% \textcolor{red}{We should describe this not from the experiment perspective, but from the coverage perspective.} 
We divide our attacks into two setups based on the real-world feasibility of the attack, namely the \textit{image domain} and the \textit{network domain}. In the image domain, the scratch values are restricted to lie in the dynamic range of images; that is, the pixel values of the adversarially scratched image $x'$ values must lie in the range $[0,1]$ for RGB images. We initially attempted to keep the mask $m$ fixed but found that this method would often get stuck in local minima, following which we chose to evolve $m$ via Differential Evolution.
% \gi{isn't it that $x'$ should lie inthe image range, not the scratch?}.
% \lr{We defined an image $x$ to be within $[0,1]$, while here we allow pixels to go beyond 1. Maybe we should mention the simple mapping from $[0,255]$ to $[0,1]$} . This allows us to evaluate our attacks in a more realistic adversarial scenario. 
The network domain imposes no such restrictions on the scratches and the adversarial scratch values can exceed the image range.
% \gi{and adversarial examples $x'$ might exceed the image range}

We generate image-domain scratches through Differential Evolution due to its ability to specify input constraints such as scratch location and pixel values. We obtain network-domain scratches through the CMA-ES algorithm due to its ability to find solutions effectively in an unconstrained search space and through Differential Evolution by relaxing the constraints on the dynamic range of pixel values.

% \begin{itemize}
%     \item \textbf{Straight Lines.} Scratches in this format are generated by only modifying the pixels specified by the 2-D line segment joining the $(x_{0}, y_{0})$ and  $(x_{1}, y_{1})$ points on the image, with $(0,0)$ as the top-left pixel of the image. 
%     To test the effectiveness of this particular scratch type, we fix the locations of the 2D line segment, and solely evolve the pixels along the line segment. The distance metric budget $L(x,x')$ in this particular experimental setting is fixed to be strictly less than $5\%$ of pixels of the entire image. Thus, for a scratch covering $m$ pixels on the image, the adversarial scratch values can be written as a vector of $n = 3m$ (for each RGB channel) that are evolved according to a genetic algorithm.
    % \item \textbf{Bezier Curves.} Scratches in this format are generated by modifying the pixels specified by a Bezier Curve, which is a parametric curve commonly used in computer graphics and robotics. In this work our scratches are in the form of quadratic Bezier curves, which utilizes $3$ points  $P_0 = (x_{0}, y_{0})$,  $P_1 = (x_{1}, y_{1})$ and  $P_2 = (x_{2}, y_{2})$ with a weighting coefficient $t$ to draw a curve according to the following equation:
    % \begin{equation}
    %     B(t) = (1-t)^{2} P_0 + 2(1-t)tP_1 + t^2 P_2
    % \end{equation}
%     % Add figure here
% \end{itemize}

\subsection{Image-Domain Attack}
\label{sec:image-domain-attack}
% \gi{\sout{But the scratch model is related to both Image-Domain and Network Domain? If yes, it would be better to move Bezier curves out of the Image domain attack?} MJ: I ran experiments on straight lines as well for image domain.}

In the image-domain attack, individual scratches are evolved according to Differential Evolution. 
%
% We initially unsuccessfully attempted to generate scratches with fixed scratch locations. 
Our initial attempts to generate scratches with fixed scratch locations were unsuccessful. These experiments are described in-depth in Section~\ref{sec:discussion}.

Each scratch in the image-domain can be represented as line segment or as a second-order rational B\a'ezier curve~\cite{Zingl2012ARA}, which is a parametric curve commonly used in computer graphics and robotics. A B\a'ezier curve uses $3$ anchor points  $P_0 = (x_{0}, y_{0})$,  $P_1 = (x_{1}, y_{1})$ and $P_2 = (x_{2}, y_{2})$ with a weighting coefficient $W$ to draw a curve according to the following equation:
\begin{equation}
    B(t) = \frac{(1-t)^2 P_0 + 2(1-t)t W P_1 + t^2 P_2}{(1-t)^2 + 2(1-t)t W + t^2}
\end{equation}

% The parabolic curve is generated through intermediate points $Q_{0}(t)$ and $Q_{1}(t)$ as follows (Figure~\ref{fig:bezier}):
% \begin{itemize}
%     \item $Q_{0}(t)$ is varied by generating equally spaced intervals between $P_{0}$ and $P_1$.
%     \item $Q_{1}(t)$ is generated by generating equally spaced intervals between $P_{1}$ and $P_{2}$ \gi{$Q_{1}(t)$ covers equally spaced intervals between P1 and P2}. 
%     \item The parabolic curve described by $B$ is generated from the linear interpolation between $Q_{0}(t)$ and $Q_{1}(t)$.
% \end{itemize}

% \gi{figure 4 includes $Q_0$ and $Q_1$, which are not discussed, isn't it? Can we remove these?}

\begin{figure}[h!]
    \centering
    \includegraphics[width=0.4\textwidth]{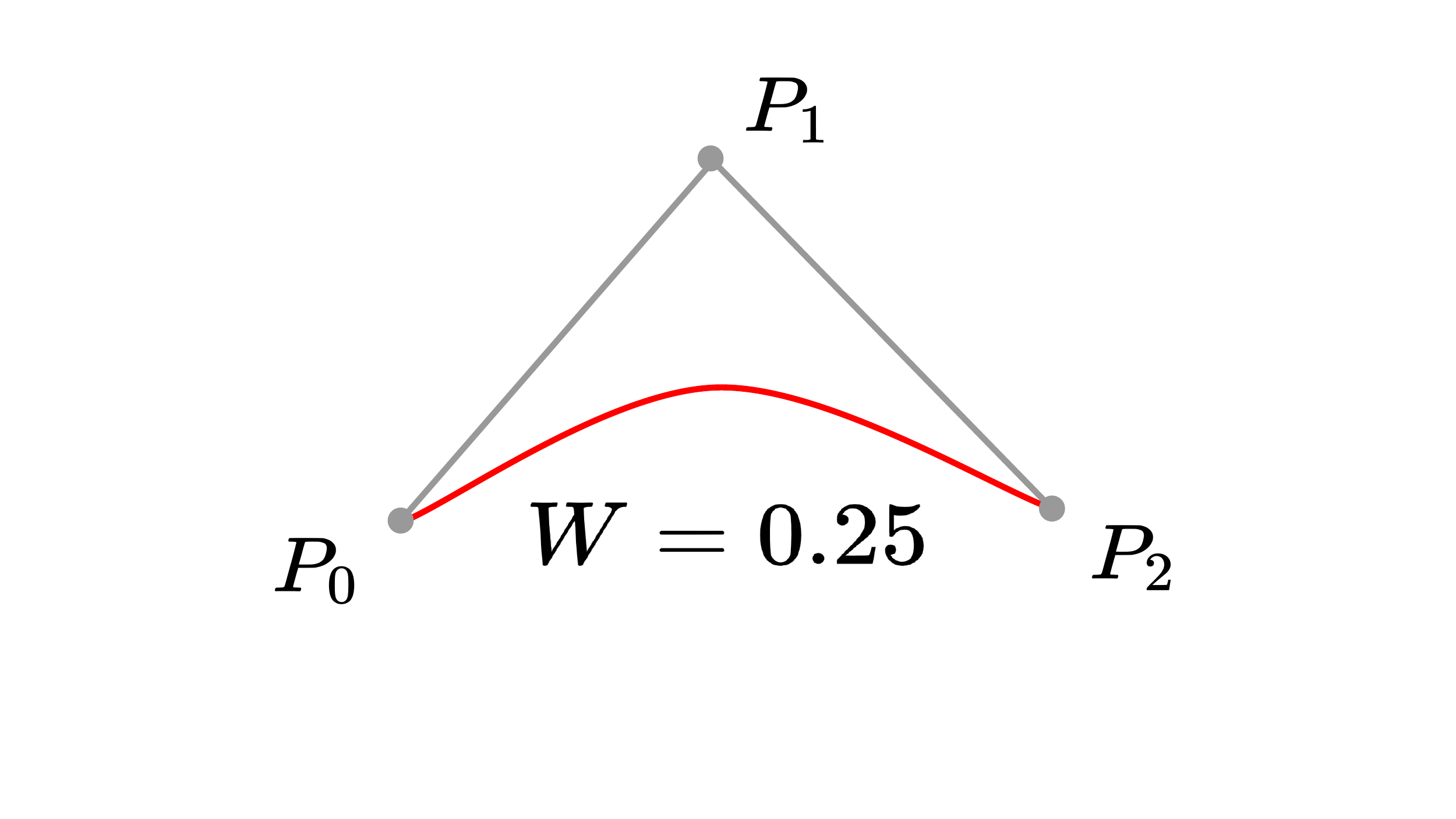}
    \caption{Second-degree \bz~curve with anchor points and weighting coefficient used to generate scratches.}
    \label{fig:bezier}
\end{figure}

% \gi{In image-domain attack we encode the perturbation..}
For image-domain attacks, the perturbation parameters are encoded into an array, which is also termed the candidate solution. For single-scratch B\a'ezier curve attacks, the candidate solution is a $10-$dimensional vector consisting of the following:

\begin{itemize}
    \item $(x_{0}, y_{0}), (x_{1}, y_{1}), (x_{2}, y_{2})$ for $P_0, P_1$, and $P_2$ respectively,  namely the x-y coordinates for the anchor points 
    % \lr{Is 'anchor points point' a typo?}. 
    Each $(x,y)$ pair is constrained to be between $[0, w], [0,h]$ so as to be within the image dimensions.
    % \item $P_1$ consisting of $(x_{1}, y_{1})$, the x-y coordinates for the first point. Constrained to be between $[0, w], [0,h]$.
    % \item $P_2$ consisting of $(x_{2}, y_{2})$, the x-y coordinates for the first point. Constrained to be between $[0, w], [0,h]$.
    \item $W$, the weighting coefficient used to control the shape of the curve. Constrained between $[0,7]$.
    % \item \gi{I think it is enough to mention the coordinates of $P_0$, $P_1$, $P_2$, that are constrained to belong to $[0, w], [0,h]$}
    \item $R,G,B$ consisting of the pixel values for each of the color channels of the adversarial scratch $\delta$. For image domains, they are constrained between $[0, 1]$. 
    % \gi{that should be $\delta$, right?}
\end{itemize}

For line segments, the scratches are encoded as a $7-$dimensional vector consisting of the following:

\begin{itemize}
    \item $(x_{0}, y_{0}), (x_{1}, y_{1})$ for $P_{0}$ and $P_{1}$, the x-y coordinates for the starting and ending points of the line segment. Each $(x,y)$ pair is constrained between $[0, w], [0,c]$.
    % \item $P_1$ consisting of $(x_{1}, y_{1})$, the x-y coordinates for the first point. Constrained to be between $[0, w], [0,c]$.
    \item $R,G,B$ consisting of the pixel values for each of the color channels of the adversarials scratch $\delta$. For image domains, they are constrained between $[0, 1]$. 
\end{itemize}

% \gc{Can we say something about these constraints, more intuition? How are the values defined? The [0,1] is obvious. c is also used as a source class.}

Multiple scratches consist of copies of the same set of variables encoded in the same array. For example, an array for 2 B\a'ezier curve scratches would be $20-$dimensional, with the first $10$ devoted to the first scratch and the next $10$ devoted to the second scratch.

\noindent\textbf{Fitness Function.} \label{fitness}
We outline the algorithm to generate targeted attacks as follows. Formally, let $F: \mathbb{R}^{d} \rightarrow \mathbb{R}$  \label{eq:1} be the fitness function that must be minimized for differential evolution, with the image classifier $f$ represented as $f: [0,1]^{w.h.c} \rightarrow \mathbb{R}^{K}$. 
% \gi{It should be better to define the classifier like this in the problem formulation, introducing the number of classes $K$}. 
The fitness function takes in the candidate solution as a vector of real numbers and produces a real number as output, which indicates the fitness of the given candidate solution. A reasonable expectation for the fitness function is to directly use the target class softmax probability; however, we observed that it is more efficient and effective to jointly minimize the top class, while attempting to maximize the target class probabilities, in addition to using their logarithms for numerical stability. Specifically, for a target class $t$ with source class $s$, we maximize the following to steer the network output towards class $t$:
\begin{equation} \label{eq:img_domain_targeted}
    F(x) = \alpha \times  \mathrm{log}\left(f(x)_{t}\right) - \beta \times \mathrm{log}\left(f(x)_{s}\right)\,,
\end{equation}
% \gi{to steer the network output towards class $c$}.

After performing a grid search, we observe that the parameters $\alpha=1, \beta=50$ offer the best attack success rate. To generate the adversarial sample $x'$, we used this fitness function in an iterative manner. This fitness function was then used to evaluate candidates at each generation in Algorithm~\ref{algorithm:differential_evolution}.

For untargeted attacks, we modify the fitness function as follows:
\begin{equation} \label{eq:img_domain_untargeted}
    F(x) = - \sum_{i=0}^{K} f(x)_i \times \mathrm{log}\left(f(x)_{i}\right)
\end{equation}
as this would maximize the entropy of the classifier predictions rather than the prediction on any particular class. 

% Let $s \in \mathbb{R}^{d}$ denote the candidate solution in the population.
% Write algorithm here to generate solutions
% The process then to generate a single adversarial scratch is in Algorithm \ref{diffe_alg}.

\subsection{Network-Domain Attack}

Network-domain attacks may be evolved according to Differential Evolution or CMA-ES, and may be further divided into two types depending on whether the parameters that govern the scratch locations are fixed or variable. Unlike fixed-scratch attacks in the image domain, we were able to obtain successful results for this setting. We explore two options that control the locations for each scratch.
% \gi{It would be better ot provide some argument why we did not consider fixed scratch location also in the Image-Domain? It is because the Network-Domain attacks are known to be more powerful? Maybe it would be more natural to mention the variable first (as this conform with the previous section), and then discuss the fixed as a specific study..}

\noindent \textbf{Fixed Scratch Location.}
In the network-domain attack, scratches are either represented as a line segment parametrized by starting and ending points $P_0 = (x_{0}, y_{0})$ and $P_1 = (x_{1}, y_{1})$, or as a second-order B\a'ezier curve with points $P_0 = (x_{0}, y_{0}), P_1 = (x_{1}, y_{1}),$ and
$P_2 = (x_{2}, y_{2})$ with a weighting factor $W$, similar to the image-domain scenario. 
% \gi{let freely evolve the value of each pixel} 
We randomly choose a mask $m$ that stays fixed throughout the attack, and evolve the value of each pixel belonging in the scratch rather than assigning a single color to each scratch as in the image-domain attacks.
Thus, for a scratch covering $M$ pixels on a RGB image, the adversarial scratch values can be written as a vector with $3M$ elements. This leads to a much larger search space for which we use the Covariance Matrix Adaptation Evolutionary Strategy (CMA-ES). We settled upon CMA-ES due to its ability to find solutions in larger and unconstrained environments, which make it suitable for network-domain attacks.
% \gc{we need to be more specific on the extensive evaluation}
We utilized the same fitness function used for the Image Domain. 

\noindent\textbf{Variable Scratch Location.} \label{variable_loc_img_domain}
In the variable scratch location, we repeated the experiments of the image-domain experiment, but relaxed the constraint on the pixel value range, allowing it to have any value, rather than being restricted to $[0,1]$.

% \label{sec:methodology}

% !TEX root = usenix2019_v3.1.tex
\section{Experimental Setup}
\label{sec:experimental_setup}

We describe the experimental setup for our scratches on the ResNet-50 and VGG-16 neural networks trained on CIFAR-10 and pre-trained ImageNet ResNet-50 and AlexNet networks, considering both image-domain and network-domain attacks. 

\subsection{Image-Domain Setup}

We trained ResNet-50~\cite{he2016deep} and VGG-16~\cite{simonyan2014very} models on the CIFAR-10~\cite{cifar10} dataset and achieved test accuracies of $93.91\%$ and $85.77\%$, respectively. In addition to attacking these models for CIFAR-10, we attacked ImageNet pre-trained AlexNet~\cite{krizhevsky2012imagenet} and ResNet-50 models. All attacks were performed on correctly classified images.

For CIFAR-10 targeted attacks, we utilized the fitness function outlined in Equation~\ref{eq:img_domain_targeted}. We observed that CIFAR-10 attacks are most effective with a population size of $50$, mutation rate $0.8$, $50$ iterations, and cross-over rate $0.7$. We observed that targeted attacks on ImageNet trained models failed to converge after several hundred iterations; hence, we proceeded with the untargeted fitness function, outlined in Equation~\ref{eq:img_domain_untargeted}, with a population size of $100$ over $100$ iterations, based on a grid search on potential populations and iterations. All CIFAR-10 experiments were conducted on $1000$ test-set images, and ImageNet experiments were conducted on $100$ validation set images.

\subsection{Network Domain Setup}
\label{sec:network-domain-setup}
\noindent\textbf{Fixed Scratch Location.}
We conducted our network-domain experiments on the ResNet50 model \cite{he2016deep}, with testset accuracy of 93.91\% and 76.13\% (top-1 accuracy) for CIFAR10 and ImageNet, respectively. All the attacks were performed on $100$ images for ImageNet and $1000$ for CIFAR-10 taken from the validation set. Selected images were originally correctly classified by the model.

Each image evaluation involved a different number of adversarial scratches whose pixel values were optimized with the CMA-ES algorithm. The algorithm was initialized with parameters $\mu = 0$ and $\sigma = 0.5$, and a population size $\lambda~=~40$, chosen after hyperparameter tuning (described in Section \ref{sec:fixed-location-network-domain}). For CIFAR-10 and ImageNet attacks, the maximum number of iterations were set to 40 and 400, respectively, for a total of $16,000$ and $160,000$ maximum queries, respectively.

All network-domain attacks were targeted. Targets were chosen at random and were fixed for each source image to make fair comparisons between attacks on the same image with different numbers of scratches.
To reduce the effects of lucky initialization of the location of the scratches, each image was attacked $10$ times with different random scratches locations chosen each time.
% The sentence above needs a refactoring, feel free to modify it and in that case delete this comment

% \noindent\textbf{Variable Scratch Location.}
% \gc{This section will not exist anymore - we need to revisit this paragraph}
% The experiments in Section~\ref{variable_loc_img_domain} were repeated with modified constraints on the pixel values. The pixel values were allowed to have the dynamic range $[-15, 15]$ to enable a large variation in potential pixel values, while still achieving a bounded solution space for our differential evolution solver.

% !TEX root = usenix2019_v3.1.tex
\section{Results}
\label{sec:results}

We evaluated our scratches on the ResNet-50 and VGG-16 neural networks trained on CIFAR-10 and pre-trained ImageNet ResNet-50 and AlexNet networks. For each dataset, we evaluated both image-domain and network-domain attacks across different types and numbers of scratches. 
\subsection{Image Domain}

% BH here COMMENTING THIS FOR A BIT
\begin{figure*}[!t]
\centering
% Original Classes
    \begin{subfigure}{.24\linewidth}
      \centering
      \includegraphics[width=0.75\linewidth]{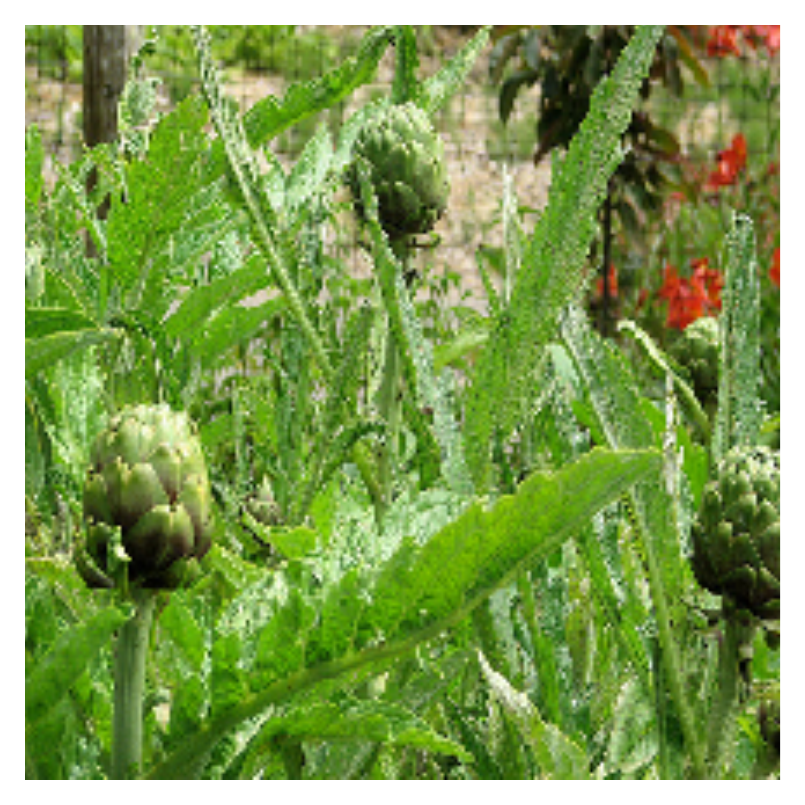}
      \vspace{-0.7em}
      \caption*{Artichoke}
    \end{subfigure}%
    \hspace{-3.5em}
    \begin{subfigure}{.24\linewidth}
      \centering
      \includegraphics[width=0.75\linewidth]{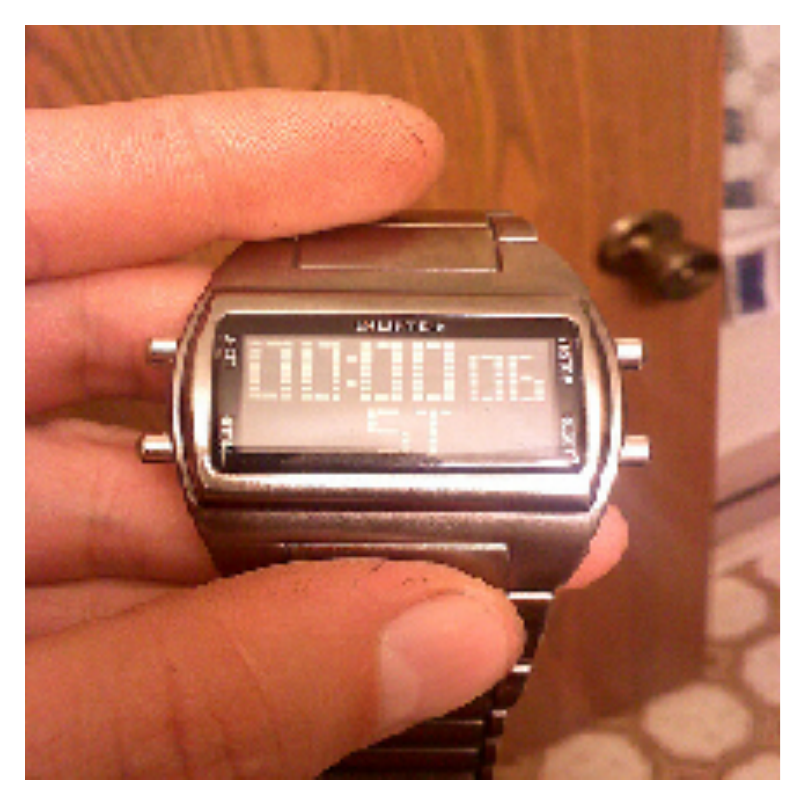}
      \vspace{-0.7em}
      \caption*{Stopwatch}
    \end{subfigure}%
    \hspace{-3.5em}
    \begin{subfigure}{.24\linewidth}
      \centering
      \includegraphics[width=0.75\linewidth]{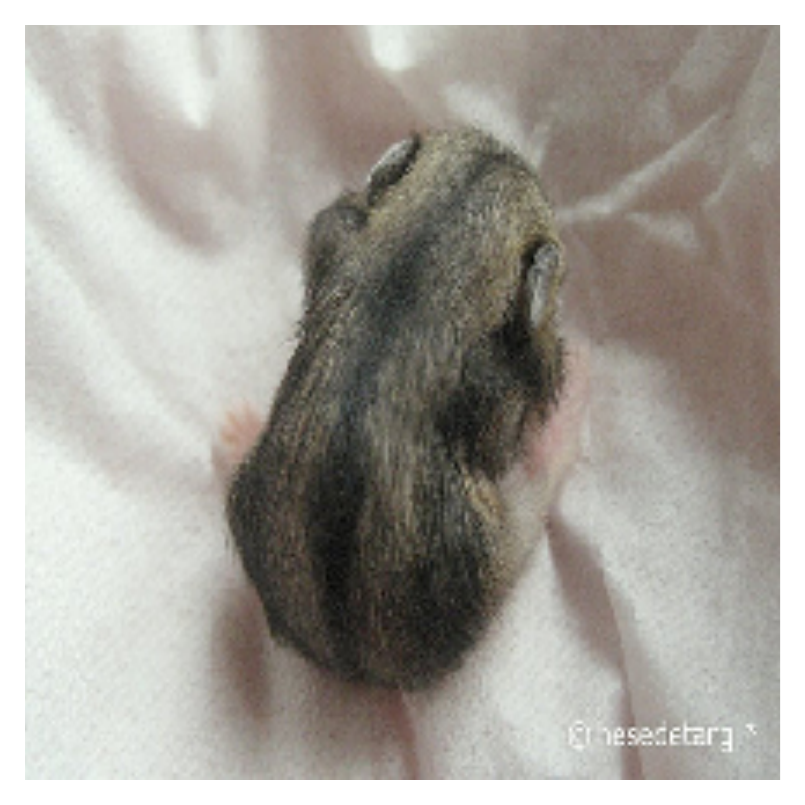}
      \vspace{-0.7em}
      \caption*{Hamster}
    \end{subfigure}%
    \hspace{-3.5em}
    \begin{subfigure}{.24\linewidth}
      \centering
      \includegraphics[width=0.75\linewidth]{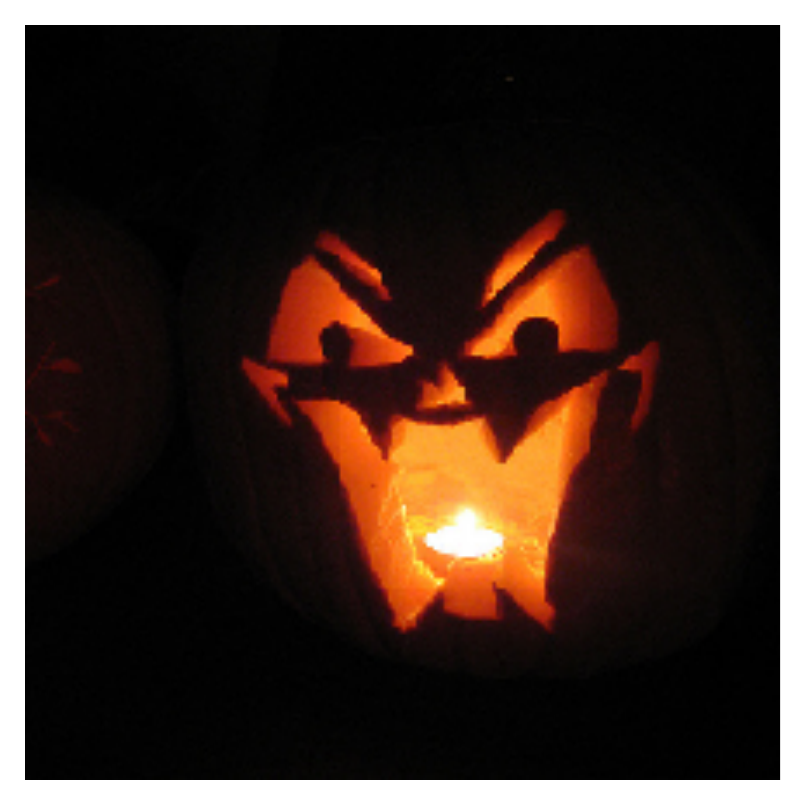}
      \vspace{-0.7em}
      \caption*{Jack-o'-lantern}
    \end{subfigure}
    \\
% Adversarial Predicted Classes
    \begin{subfigure}{.24\linewidth}
      \centering
      \includegraphics[width=0.75\linewidth]{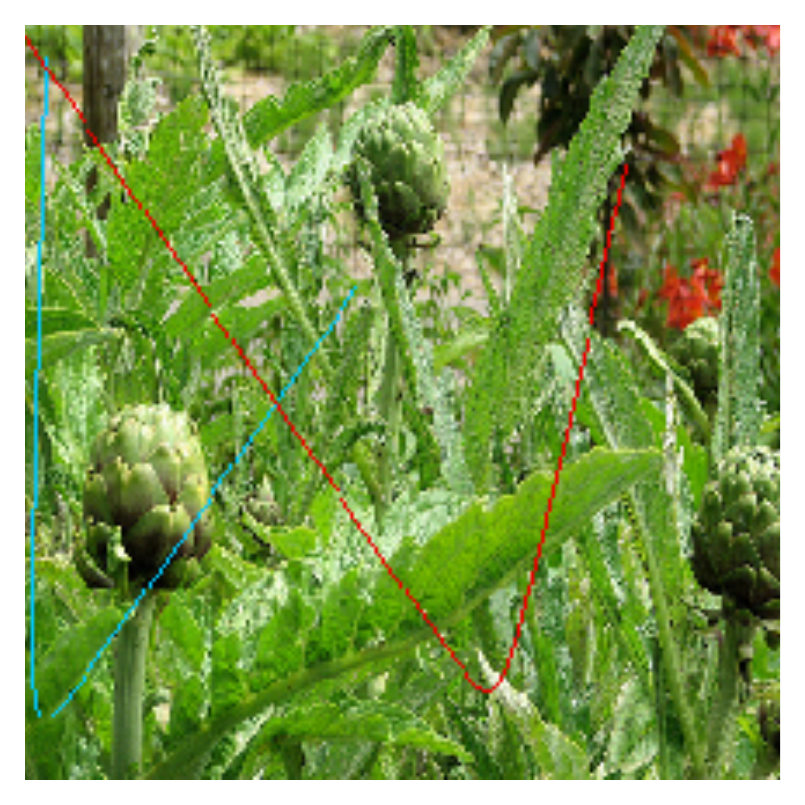}
      \vspace{-0.7em}
      \caption*{Spider's web}
    \end{subfigure}%
    \hspace{-3.5em}
    \begin{subfigure}{.24\linewidth}
      \centering
      \includegraphics[width=0.75\linewidth]{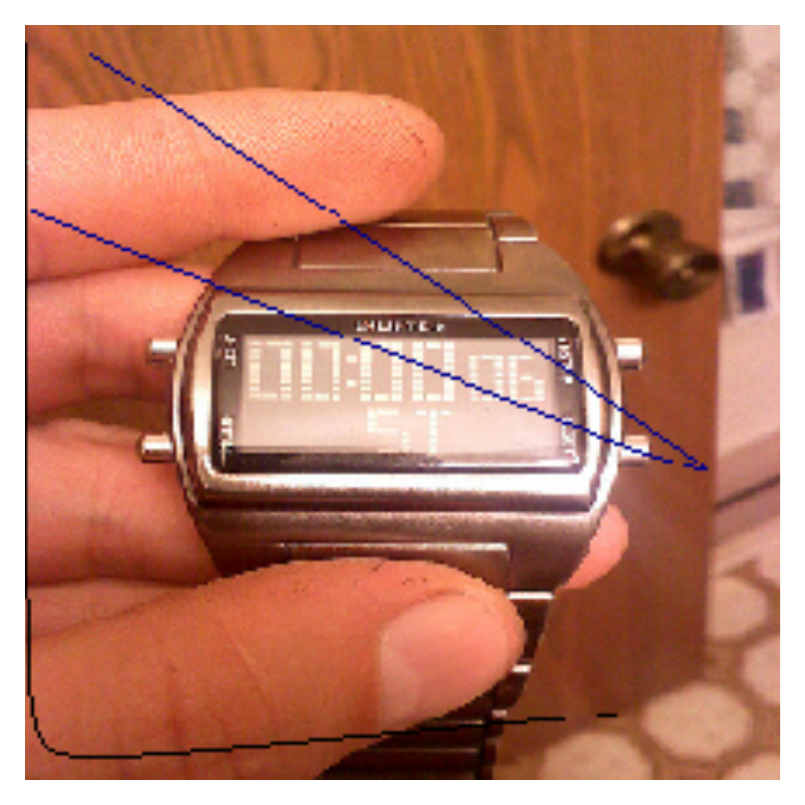}
      \vspace{-0.7em}
      \caption*{Purse}
    \end{subfigure}%
    \hspace{-3.5em}
    \begin{subfigure}{.24\linewidth}
      \centering
      \includegraphics[width=0.75\linewidth]{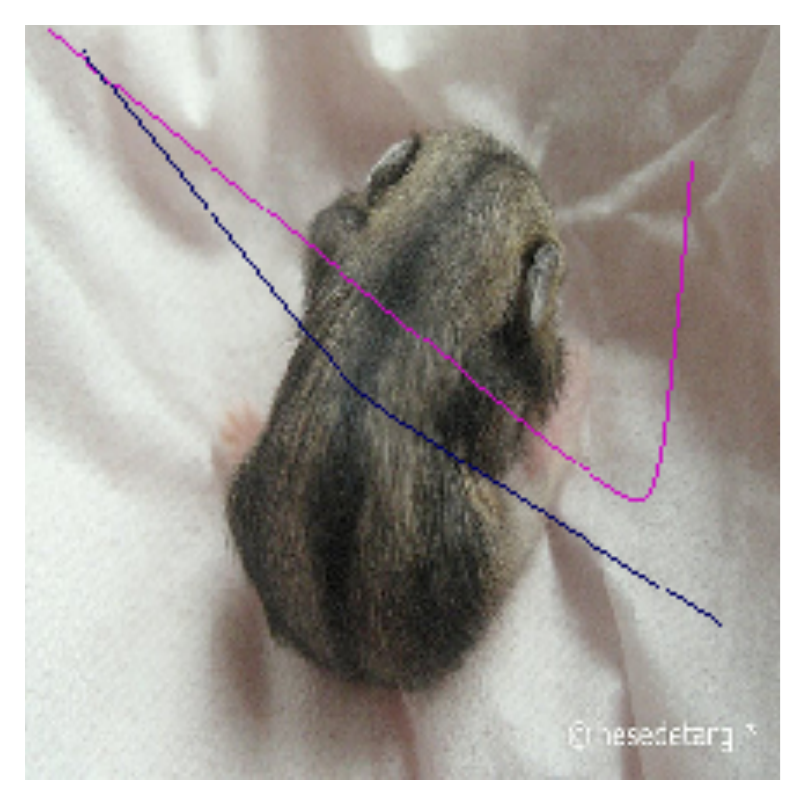}
      \vspace{-0.7em}
      \caption*{Mitten}
    \end{subfigure}%
    \hspace{-3.5em}
    \begin{subfigure}{.24\linewidth}
      \centering
      \includegraphics[width=0.75\linewidth]{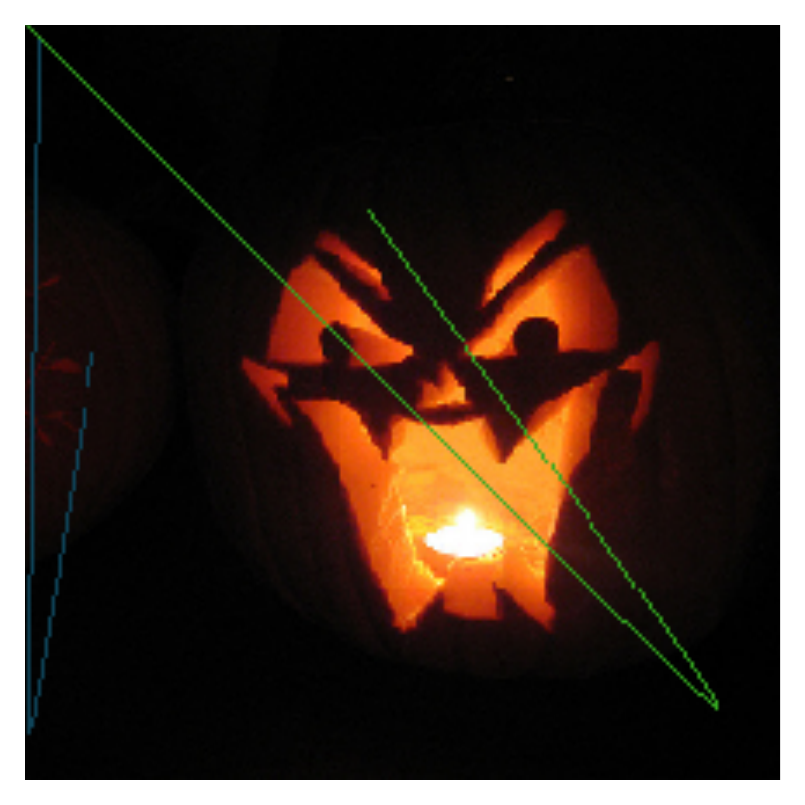}
      \vspace{-0.7em}
      \caption*{Violin}
    \end{subfigure}
    \caption{Variable location, double scratch attacks on ImageNet in the image domain. The top row contains the original images and their respective class label. The bottom row images are adversarial samples generated via our double-scratch attack and the predicted class when evaluated on ResNet-50~\cite{he2016deep}.}
    % \gc{MODEL + add reference in the text}. Best viewed in color. }
    \label{fig:img-attacks}
\end{figure*}

\noindent\textbf{Straight Lines.}
We first applied adversarial scratches in the form of a straight line, with several examples being demonstrated in Figure~\ref{fig:examples-img-attacks}. 
%
% and ~\ref{fig:tigerfig2} \lr{Shall we just refer to Figure 2? If we want to refer to specific instances of straight line attacks, then I think we should do the same for Bézier curve attacks. Also, the actual examples of straight line attacks are Figure 2b and 2c, not 2a and 2b}.
%
Table~\ref{tab:image_domain_variable_scratch_cifar10} reports the CIFAR-10 attack success rates. We note that straight line image-domain attacks were less successful on ImageNet, with $2-$straight line scratches achieving a $60\%$ success rate.

\noindent\textbf{B\a'ezier Curves.}
% \highlight{split this into straight lines and bezier curves?}
Image-domain attacks demonstrated a high success rate across targeted attacks in CIFAR-10 and untargeted attacks in ImageNet, across B\a'ezier curve and straight line shapes, and across varying numbers of scratches. We observe in Table~\ref{tab:image_domain_variable_scratch_cifar10} that more scratches resulted in higher attack success rates, queries were largely independent of the number of scratches, and B\a'ezier-curve-based attacks were more successful than straight-line-based attacks. For ImageNet-based models, we achieved a $70\%$ success rate for 2-B\a'ezier curve attacks with a population size of $500$ evaluated over $100$ iterations, and $30\%$ success rate for 2 straight lines with the same population and iterations. We observed that image-domain attacks were significantly harder to achieve than network-domain attacks, notably because of the extremely small search space both in terms of dimensionality and the constraints on the upper and lower bounds of the search space. Nevertheless, our experiments show a clear vulnerability in neural-network-based classifiers to this class of attacks. Further details can be found in Table~\ref{tab:image_domain_variable_scratch_cifar10}.

\begin{table*}[ht!]
\centering
\begin{tabular}{cccccccc}
\hline
\multicolumn{8}{c}{\textbf{Image Domain}} \\ \hline
\multicolumn{4}{c|}{\textbf{Bézier Curves}} & \multicolumn{4}{c}{\textbf{Straight Lines}} \\ \hline
\textbf{Scratches} & \textbf{Success Rate} & \textbf{Queries} & \multicolumn{1}{c|}{\textbf{Coverage}} & \textbf{Scratches} & \textbf{Success Rate} & \textbf{Queries} & \textbf{Coverage} \\ \hline
\multicolumn{1}{c|}{1} & \multicolumn{1}{c|}{66.25\%} & \multicolumn{1}{c|}{711} & \multicolumn{1}{c|}{4.13\%} & \multicolumn{1}{c|}{1} & \multicolumn{1}{c|}{61.1\%} & \multicolumn{1}{c|}{456} & 2.64\% \\
\multicolumn{1}{c|}{2} & \multicolumn{1}{c|}{86.5\%} & \multicolumn{1}{c|}{1140} & \multicolumn{1}{c|}{5.95\%} & \multicolumn{1}{c|}{2} & \multicolumn{1}{c|}{75.39\%} & \multicolumn{1}{c|}{581} & 4.96\% \\
\multicolumn{1}{c|}{3} & \multicolumn{1}{c|}{92.06\%} & \multicolumn{1}{c|}{899} & \multicolumn{1}{c|}{7.48\%} & \multicolumn{1}{c|}{3} & \multicolumn{1}{c|}{83.33\%} & \multicolumn{1}{c|}{490} & 6.82\% \\ \hline
\end{tabular}
\caption{Variable-location image-domain scratch attack results for ResNet-50 trained on CIFAR-10. More scratch coverage resulted in higher success rates, with B\a'ezier curves offering a better attack success rate than straight lines.}
\label{tab:image_domain_variable_scratch_cifar10}
\end{table*}

\begin{table*}[ht!]
\centering
\begin{tabular}{cccccccc}
\hline
\multicolumn{8}{c}{\textbf{Image Domain}} \\ \hline
\multicolumn{4}{c|}{\textbf{Bezier Curves}} & \multicolumn{4}{c}{\textbf{Straight Line}} \\ \hline
\textbf{Scratches} & \textbf{Success Rate} & \textbf{Queries} & \multicolumn{1}{c|}{\textbf{Coverage}} & \textbf{Scratches} & \textbf{Success Rate} & \textbf{Queries} & \textbf{Coverage} \\ \hline
\multicolumn{1}{c|}{2} & \multicolumn{1}{c|}{70\%} & \multicolumn{1}{c|}{30472} & \multicolumn{1}{c|}{0.7\%} & \multicolumn{1}{c|}{2} & \multicolumn{1}{c|}{33\%} & \multicolumn{1}{c|}{32198} & 0.56\% \\
\multicolumn{1}{c|}{3} & \multicolumn{1}{c|}{73\%} & \multicolumn{1}{c|}{21988} & \multicolumn{1}{c|}{1.3\%} & \multicolumn{1}{c|}{3} & \multicolumn{1}{c|}{42\%} & \multicolumn{1}{c|}{25499} & 0.9\% \\ \hline
\end{tabular}
\caption{Variable-location image-domain scratch attack results for ResNet-50 trained on ImageNet. More scratch coverage resulted in higher success rates, with B\a'ezier curves offering a better attack success rate than straight lines. ImageNet requires significantly more queries than CIFAR-10 attacks.}
\label{tab:img_domain_results}
\end{table*}

\noindent\textbf{Comparing to Existing Attacks.}
We compare our B\a'ezier-curve attack to the one-pixel attack~\cite{su2019one}, which is the most similar attack. The one-pixel attack uses differential evolution to find \textit{single} pixels that can fool classifiers. For CIFAR-10, we trained a VGG-16 model to a test-accuracy of $85.77\%$, and compared our single-scratch, image-domain B\a'ezier curve attack to theirs. We obtained an attack success rate of $75.79\%$ for a query limit of $2500$ compared to their attack success rate of $5.63\%$ with a query limit of $20,000$. Table~\ref{tab:single_pixel_comparison_vgg16} highlights further dependence on attack cost. For ImageNet-based models, we attacked the AlexNet model and were able to successfully fool $100\%$ of all images under $50,000$ queries, while the one-pixel attack achieves a success rate of $16.04\%$ with the same query limit.

\begin{table}[ht!]
\centering
\begin{tabular}{|C{1.9cm}|c|c|c|C{1.45cm}|}
\hline
                        & \multicolumn{3}{c|}{\textbf{Single Scratch attack}} & \textbf{One pixel attack} \\ \hline
\textbf{Query Limit}    & 2500           & 10000        & 20000       & 20000    \\ \hline
\textbf{Success Rate}   & 75.79\%        & 83.7\%       & 84.7\%      & 5.63\%   \\ \hline
\textbf{Pixel Coverage} & 3.78\%         & 3.77\%       & 3.68\%      & 0.1\%    \\ \hline
\end{tabular}
\caption{Our attack compared to one-pixel attack~\cite{su2019one} for VGG-16 trained on CIFAR-10.
% \lr{I don't feel like this is a strong comparison. The one-pixel attack uses only 1 pixel, while our scratches are 100x larger, so it is reasonable to have an higher success rate} 
% Cost here is measured as the product of the population size and the total number of iterations.
}
\label{tab:single_pixel_comparison_vgg16}
\end{table}

\noindent\textbf{Population Size Tuning.}
To choose an appropriate population size that optimizes for attack success rate as well as latency, we varied the population size and found that a population of $50$ offered the best performance.

\noindent\textbf{Physical Attack.}
We attempted to attack the Microsoft Cognitive Services Image Captioning API~\cite{vision-api} using our proposed methods. The service provides a descriptive caption of the image as well as its confidence. Our goal was to utilize the confidence as our fitness function and attempt to minimize the confidence to make the API generate fake or missing captions. In this setting, we used $3$ scratches, with $50$ iterations and a population size of $50$, with an upper limit of $2500$ queries to the API.

We successfully managed to decrease the caption confidence from an initial value of $0.87$ down to $0.25$, and were able to deceive the API into generating wrong captions. In one particular case, the API was not able to generate any captions at all. Several examples of these generated scratches are shown in Table~\ref{tab:microsoft-attacks}.

\begin{table*}[h]
    \centering
    % upper half
    \begin{tabular}{C{0.12\textwidth}|L{0.24\textwidth}|L{0.24\textwidth}|L{0.24\textwidth}}
        \hline
        \textbf{Input \newline Image} & 
            \includegraphics[width=\linewidth]{./figures/experiments/original.pdf} & 
            \includegraphics[width=\linewidth]{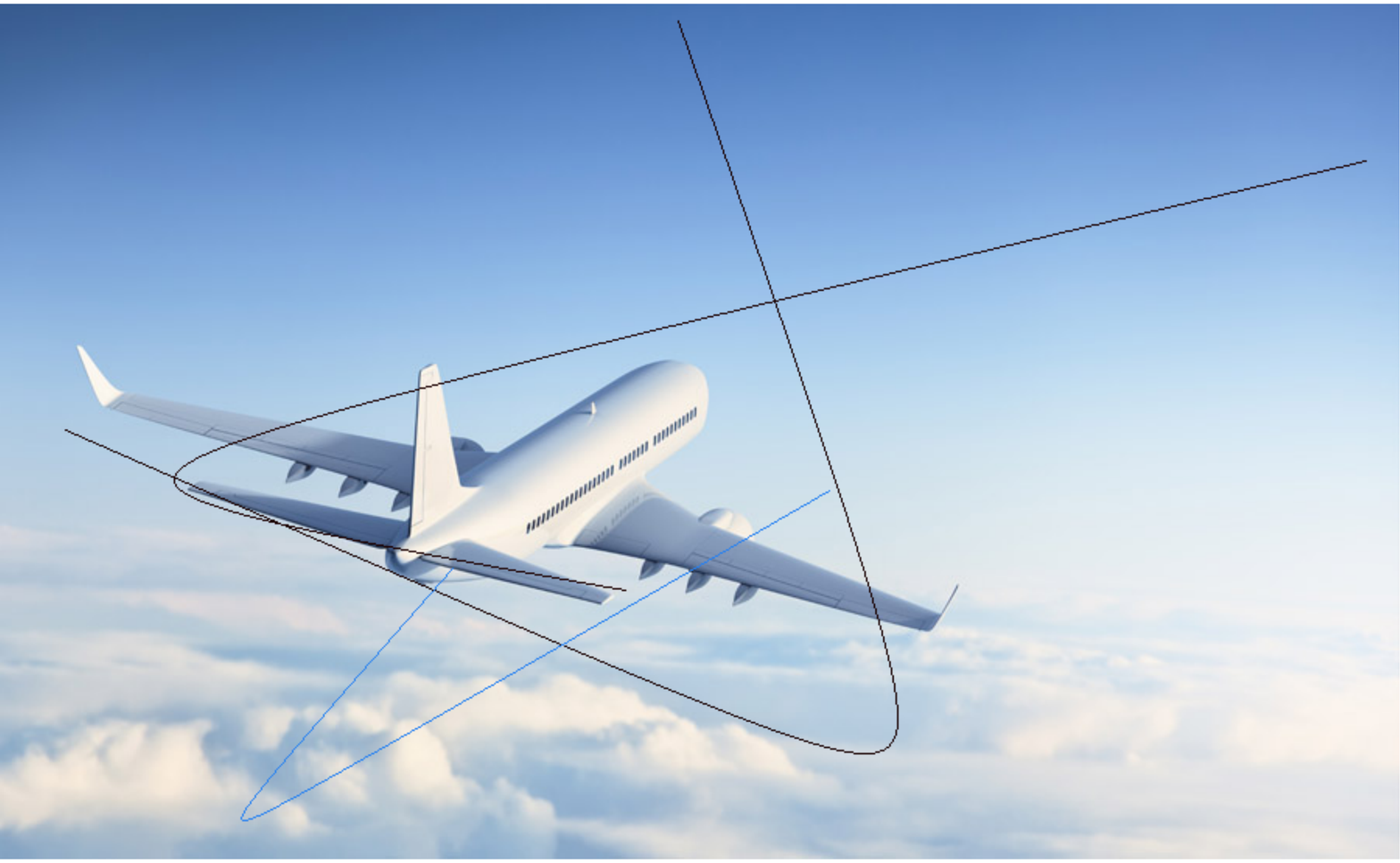} & 
            \includegraphics[width=\linewidth]{./figures/experiments/people_kites.pdf} \\ \hline
        \textbf{Predicted Caption} &
            A large passenger jet flying through a cloudy blue sky &
            A flock of birds flying in the sky &
            A group of people flying kites in the sky \\ \hline
        \textbf{Confidence} & 0.870 & 0.438 & 0.406 \\ \hline
    \end{tabular}
    % lower half
    \\[10pt] % white space between the two halves
    \begin{tabular}{C{0.12\textwidth}|L{0.24\textwidth}|L{0.24\textwidth}|L{0.24\textwidth}}
        \hline
        \textbf{Input \newline Image} & 
            \includegraphics[width=\linewidth]{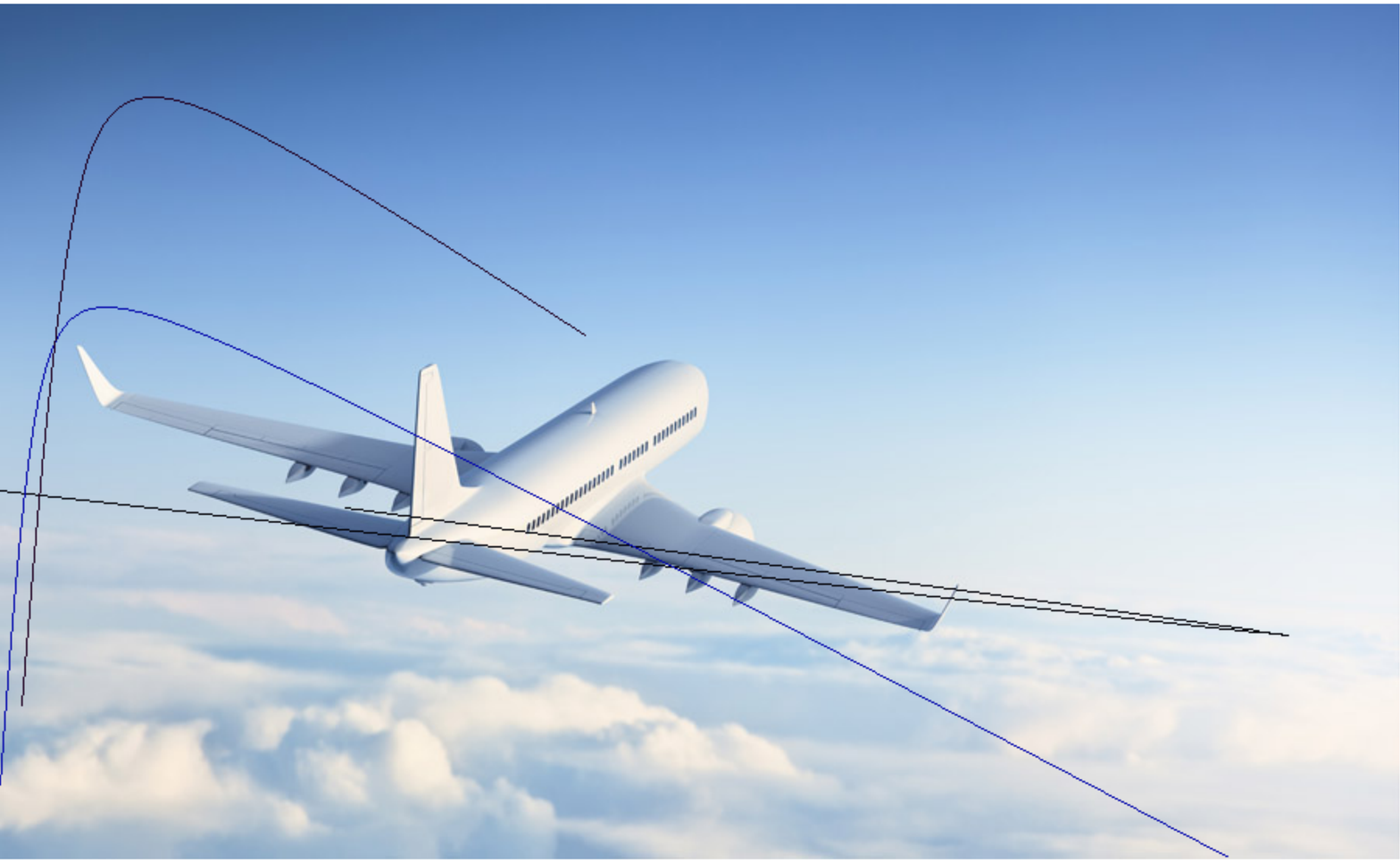}
        & 
            \includegraphics[width=\linewidth]{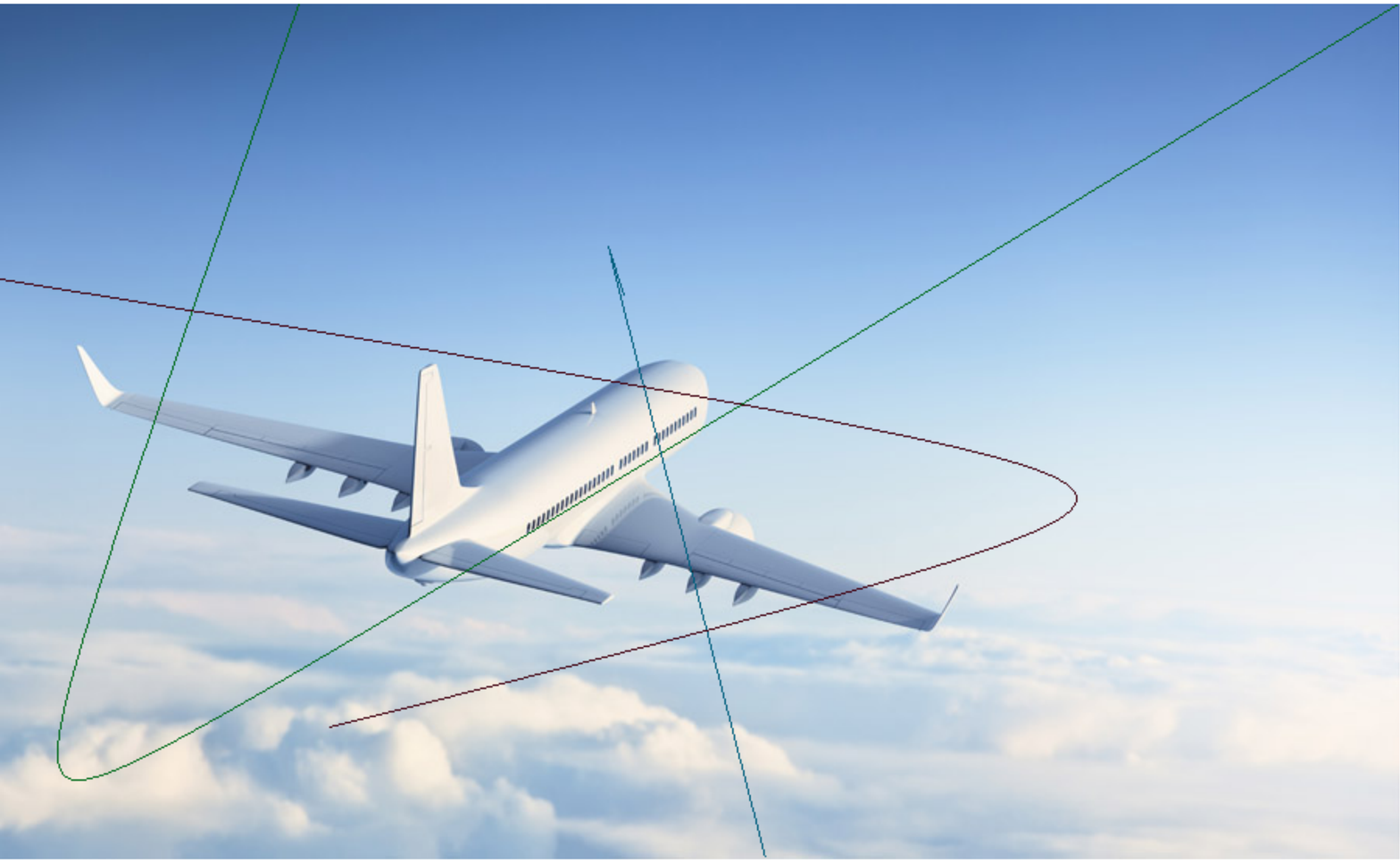}
        & 
            \includegraphics[width=\linewidth]{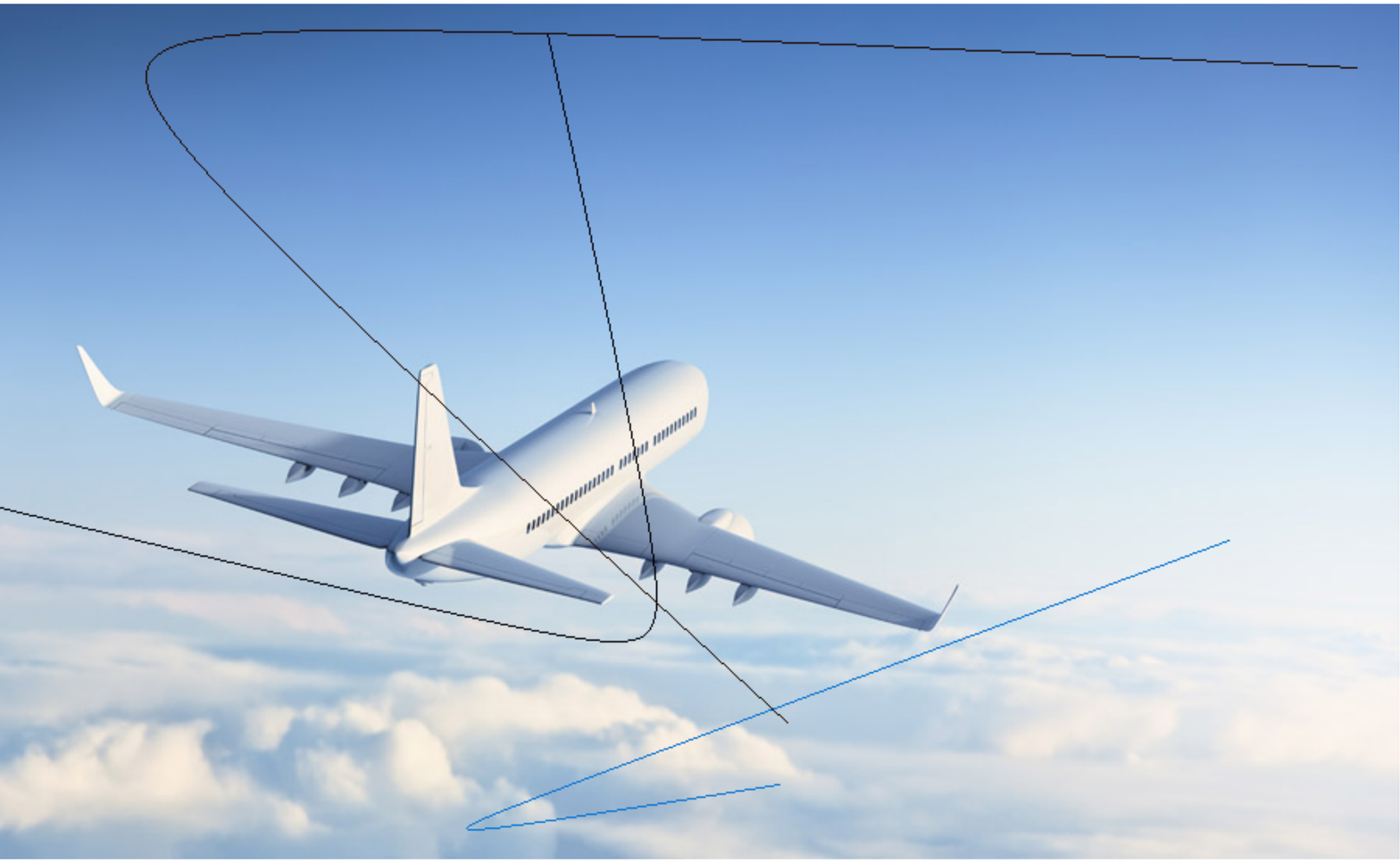}
        \\ \hline
        \textbf{Predicted Caption} &
            A flock of seagulls flying in the sky &
            A airplane that is flying in the sky &
            No caption \\ \hline
        \textbf{Confidence} & 0.317 & 0.250 & --- \\ \hline
    \end{tabular}
    % \captionsetup{width=0.7\textwidth}
    \caption{Successful attacks against the Microsoft Cognitive Services Image Captioning API. The top left image denotes the source image, which is captioned correctly as `a large passenger jet flying through a cloudy blue sky'. All other images are examples of images that successfully deceived the captioning service. Best viewed in color.}
    \label{tab:microsoft-attacks}
\end{table*}

\subsection{Fixed-Location Network Domain}
\label{sec:fixed-location-network-domain}

\begin{figure}[h!]
    \centering
    \includegraphics[clip, trim=0cm 0cm 1cm 1cm, width=\linewidth]{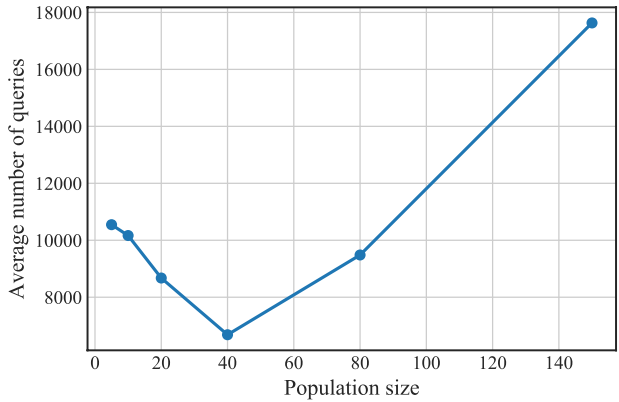}
    \caption{Impact of population size on the average number of queries needed by fixed-location network-domain attacks on CIFAR-10.} %\bh{Would it be possible to replot this? Saving it at .eps or .pdf would significantly improve it's quality!}}
    \label{fig:pop_tuning}
\end{figure}

\noindent\textbf{Population Size Tuning. }
To choose a good value for the population size, we performed attacks on a set of 24 images from CIFAR-10, each time picking a population size from $[5, 10, 20, 40, 80, 150]$. We set the population size as $40$, which we observed to give the fewest average number of queries to perform a successful attack (Figure~\ref{fig:pop_tuning}). 
% The plot in Figure \ref{fig:pop_tuning} shows that the best population size is 40, which gives the smallest average number of queries to perform a successful attack.

\begin{figure}[h]
    \centering
    \begin{subfigure}{.5\linewidth}
      \centering
    %   \footnotesize \textsf{Original Image \\ castle \\[-5pt]}
    \caption*{\textbf{Original Image}\\Castle}
    \vspace{-1.2em}
      \includegraphics[clip, trim=4cm 8cm 4cm 8cm, width=0.85\linewidth]{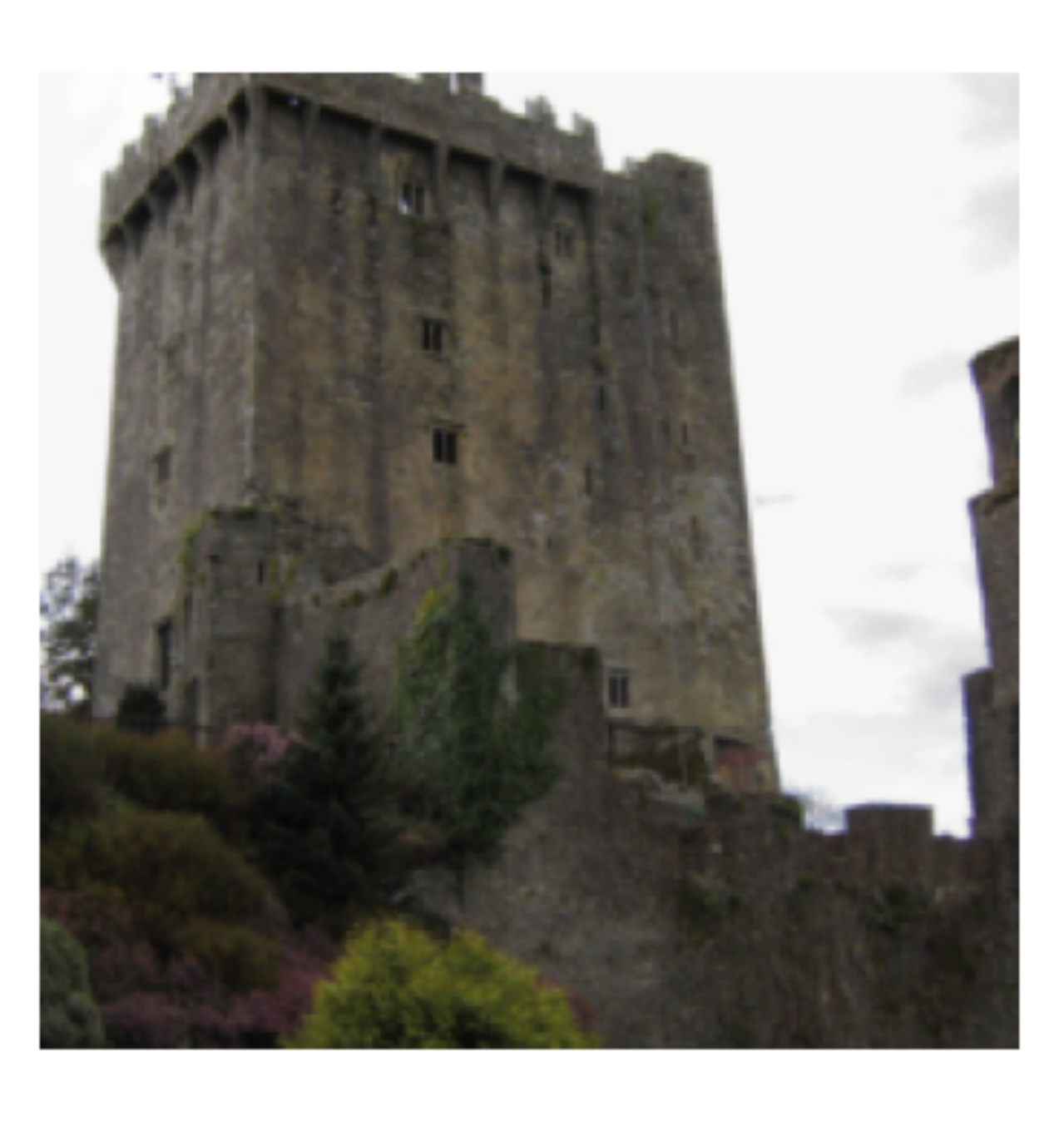}
    \end{subfigure}%
    \hspace{-2em}
    \begin{subfigure}{.5\linewidth}
      \centering
    %   \footnotesize \textsf{Adversarial Image \\ honeycomb \\[-5pt]}
      \caption*{\textbf{Adversarial Image}\\Honeycomb}
      \vspace{-1.2em}
      \includegraphics[clip, trim=4cm 8cm 4cm 8cm, width=0.85\linewidth]{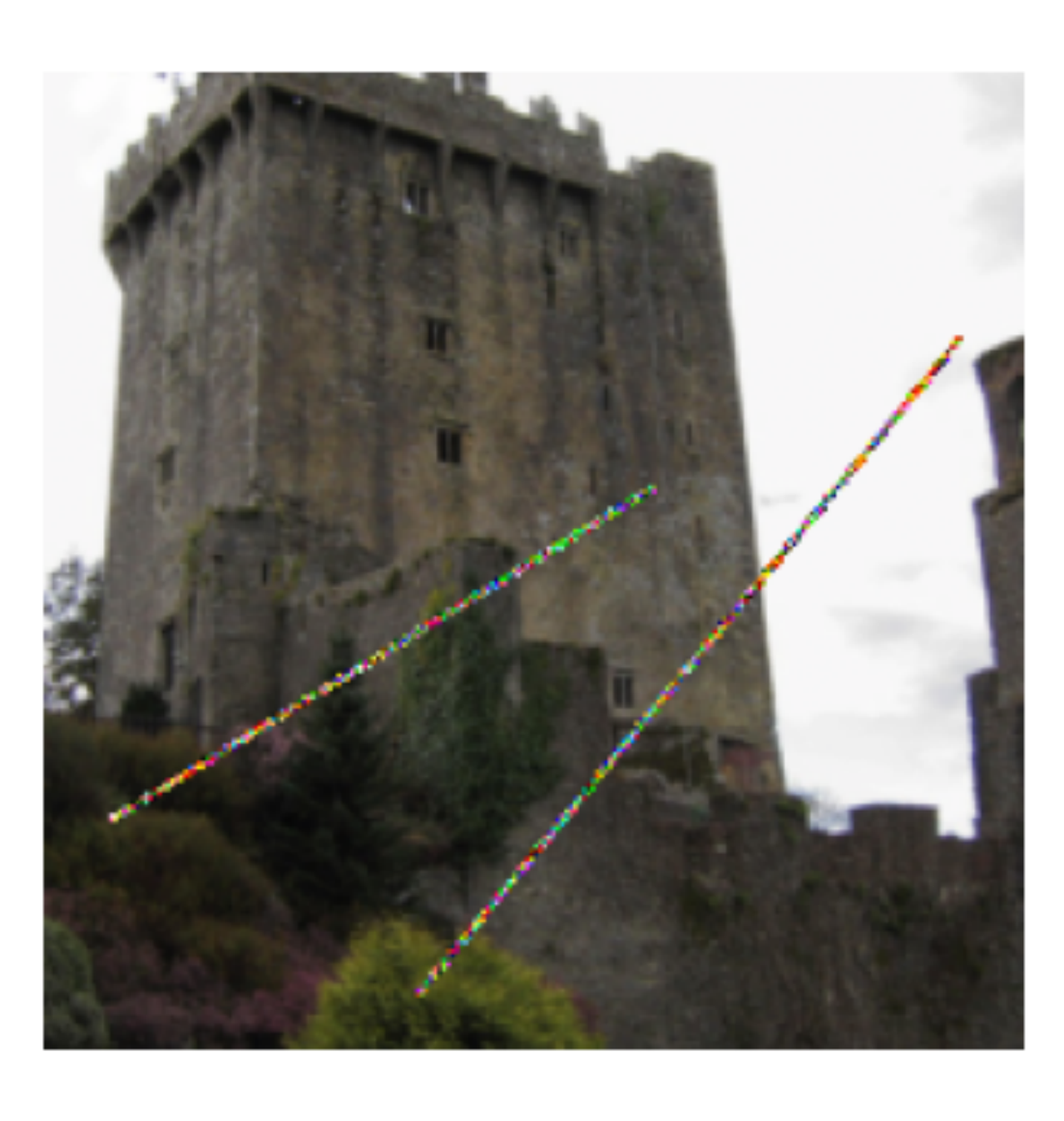}
    \end{subfigure}
    \\[5pt]
    \begin{subfigure}{.5\linewidth}
      \centering
    %   \footnotesize \textsf{Original Image \\ dugong, Dugong dugon \\[-5pt]}
      \caption*{dugong, Dugong dugon}
      \vspace{-1.2em}
      \includegraphics[clip, trim=4cm 8cm 4cm 8cm, width=0.85\linewidth]{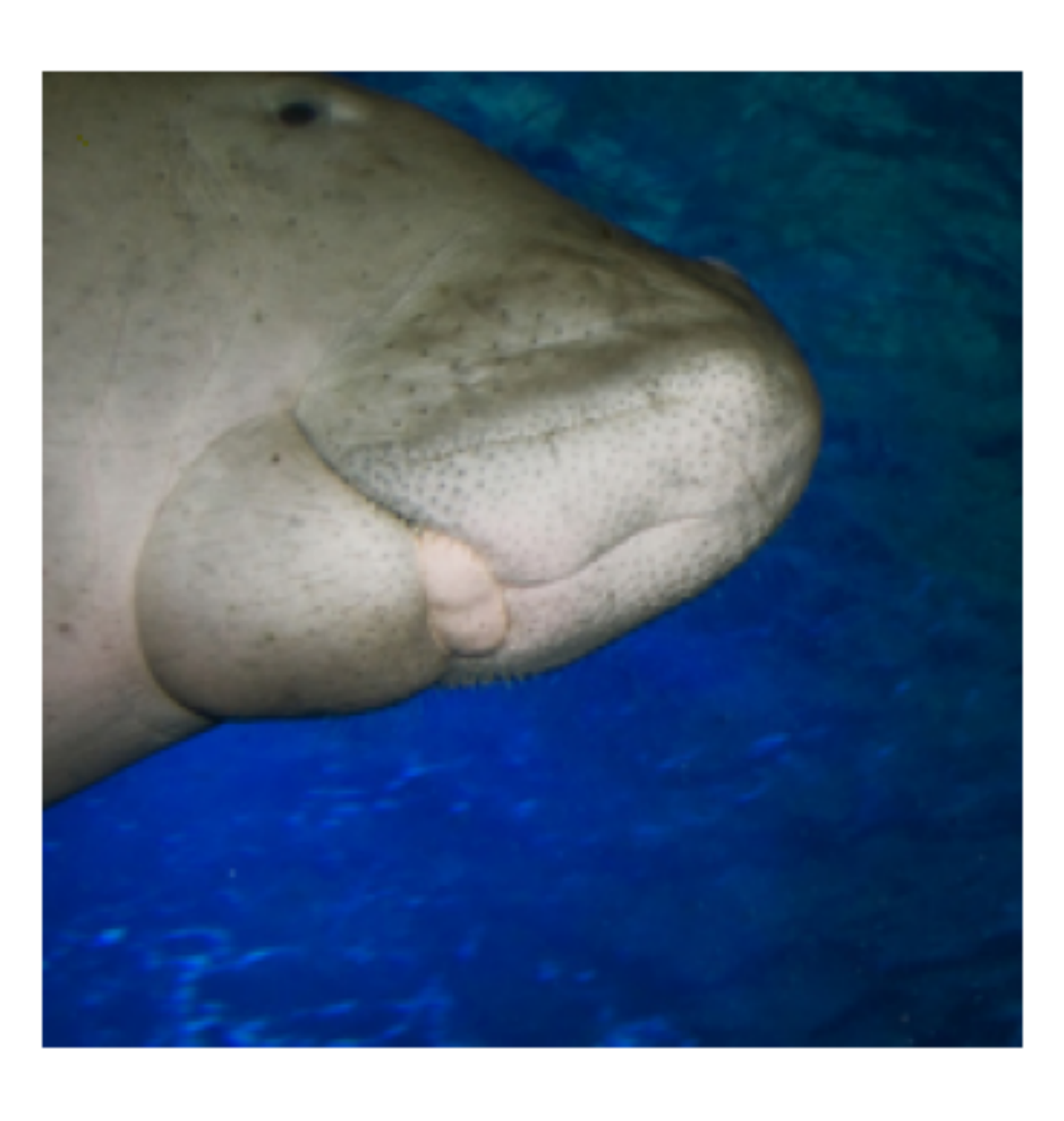}
    \end{subfigure}%
    \hspace{-2em}
    \begin{subfigure}{.5\linewidth}
      \centering
    %   \footnotesize \textsf{Adversarial Image \\ shopping cart \\[-5pt]}
      \caption*{Shopping cart}
      \vspace{-1.2em}
      \includegraphics[clip, trim=4cm 8cm 4cm 8cm, width=0.85\linewidth]{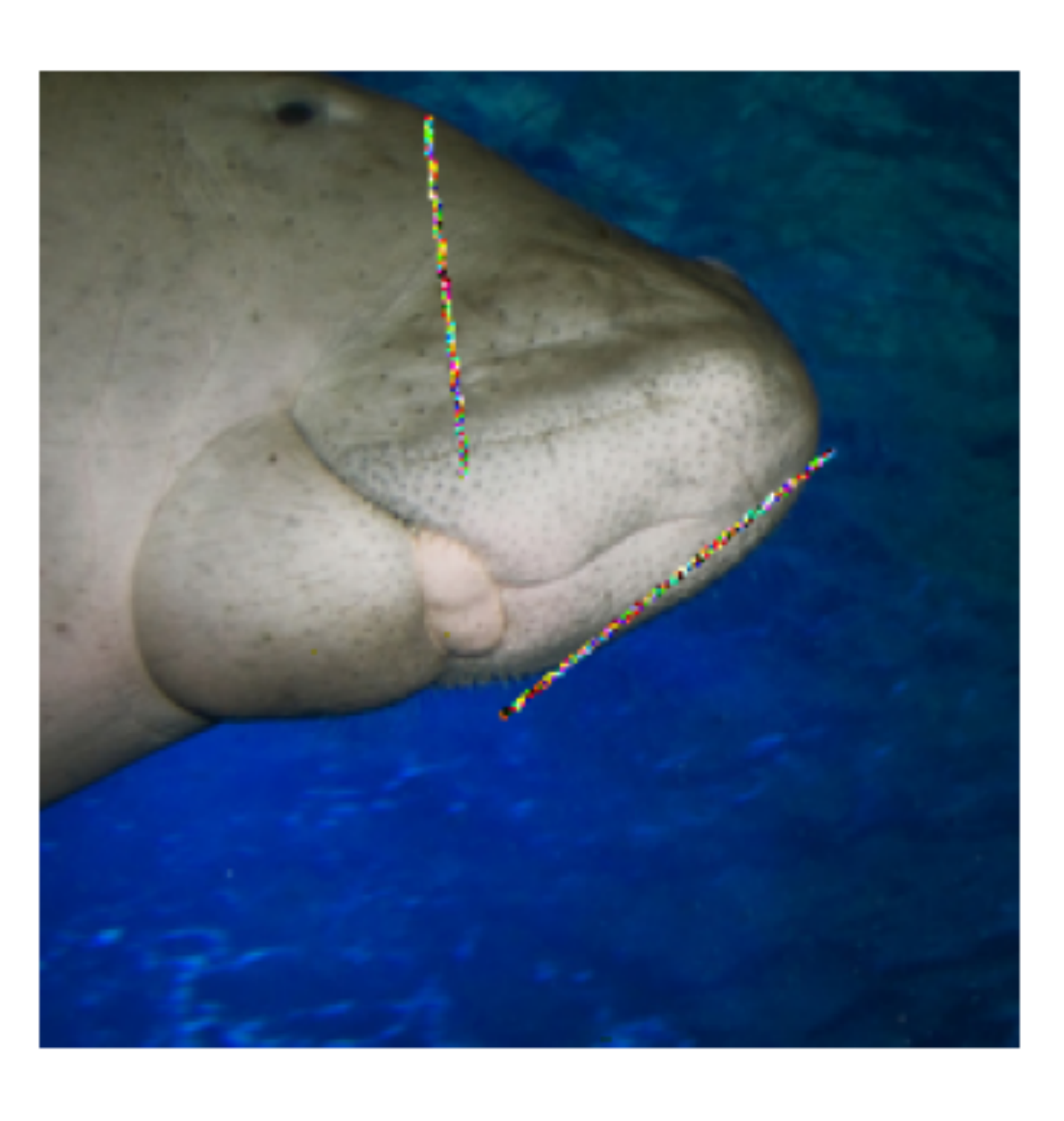}
    \end{subfigure}
    \caption{Examples of CMA-ES straight lines attacks in the network domain. Scratches were clipped to the image domain for visualization purposes. The left side shows the original images and their corresponding class; the right side shows the generated adversarial samples (scratches) and their respective predicted labels when evaluated on ResNet-50. Best viewed in color.}
    \label{fig:network-straight}
\end{figure}

\noindent\textbf{Straight lines attacks.}
%We first applied adversarial scratches with the shape of a straight line.
Figure~\ref{fig:network-straight} shows a few examples of adversarial straight line scratches; Tables~\ref{tab:network_domain_fixed_scratch_imagenet} and \ref{tab:network_domain_fixed_scratch_cifar10} show the results from 100 attacked images. We report the average percentage of pixels covered by successful attacks, the average number of queries of successful attacks, and the success rates.  It is evident, particularly in ImageNet results, that using more scratches leads to a lower average number of queries and a higher success rate.

% \begin{table}[h]
%     \centering
%     \begin{tabular}{| c | C{2cm} | c | c |}
%         \hline
%         Scratches & Mean pixels covered & Queries & Success rate \\ \hline
%         1 & 3.28\% & 3133 & 54.90\% \\ \hline
%         2 & 3.96\% & 3207 & 76.80\% \\ \hline
%         3 & 4.30\% & 2829 & 83.80\% \\ \hline
%         4 & 4.40\% & 2622 & 90.70\% \\ \hline
%     \end{tabular}
%     \caption{CIFAR10, Straight lines attacks}
%     \label{table:cifar10-straight}
% \end{table}

% \begin{table}[h]
%     \centering
%     \begin{tabular}{| c | C{2cm} | c | c |}
%         \hline
%         Scratches & Mean pixels covered & Queries & Success rate \\ \hline
%         1 & 0.56\% & 34623 & 44.00\% \\ \hline
%         2 & 0.99\% & 30776 & 80.50\% \\ \hline
%         3 & 1.42\% & 23826 & 92.40\% \\ \hline
%         4 & 1.86\% & 18848 & 96.40\% \\ \hline
%     \end{tabular}
%     \caption{ImageNet, Straight lines attacks}
%     \label{table:imagenet-straight}
% \end{table}

\begin{figure}[h]
    \centering
    \begin{subfigure}{.5\linewidth}
      \centering
    %   \footnotesize \textsf{Original Image \\ spoonbill \\[-5pt]}
      \caption*{\textbf{Original Image}\\Spoonbill}
      \vspace{-1.2em}
      \includegraphics[clip, trim=4cm 8cm 4cm 8cm, width=0.85\linewidth]{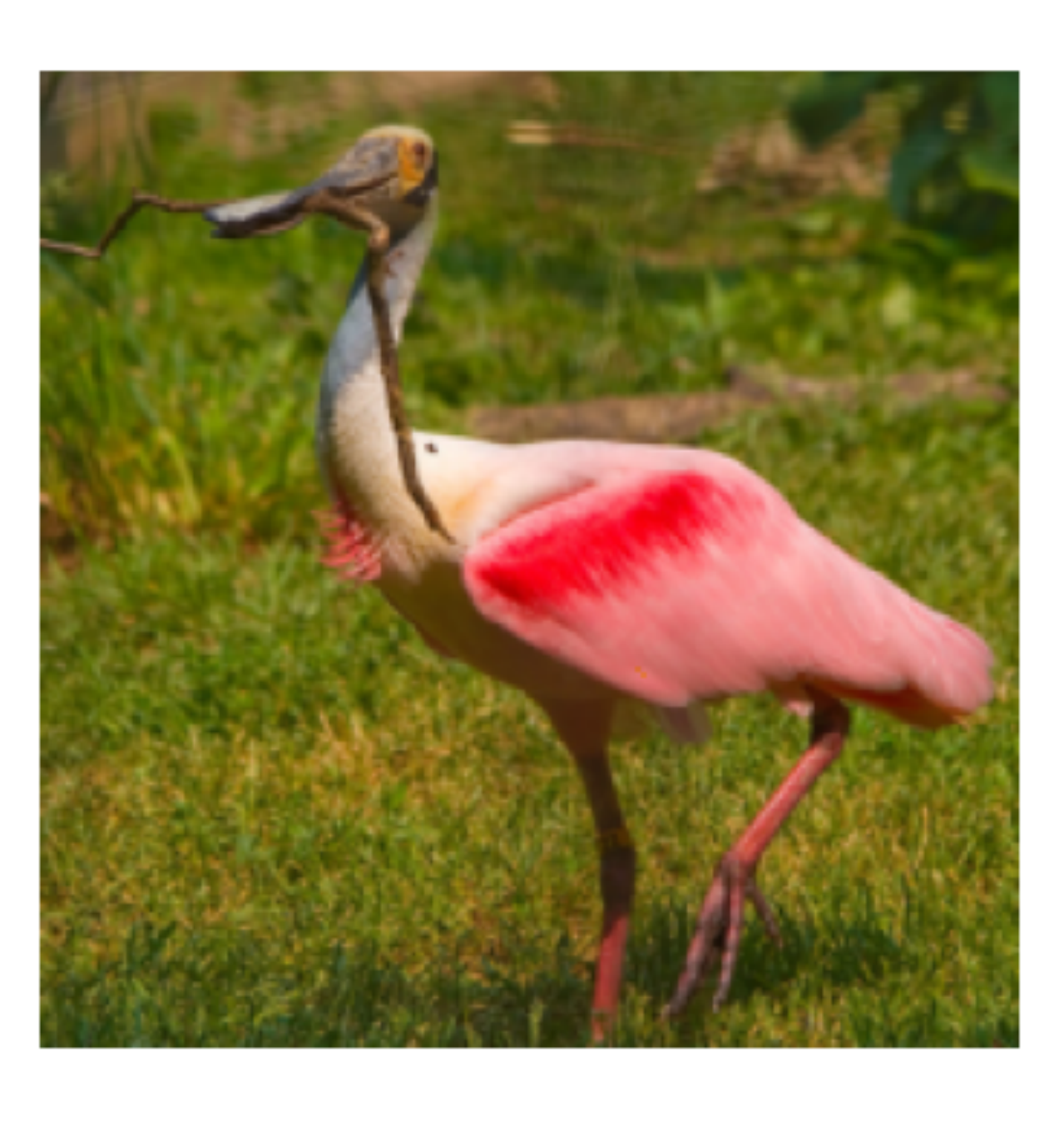}
    \end{subfigure}%
    \hspace{-2em}
    \begin{subfigure}{.5\linewidth}
      \centering
    %   \footnotesize \textsf{Adversarial Image \\ digital clock \\[-5pt]}
      \caption*{\textbf{Adversarial Image}\\Digital clock}
      \vspace{-1.2em}
      \includegraphics[clip, trim=4cm 8cm 4cm 8cm, width=0.85\linewidth]{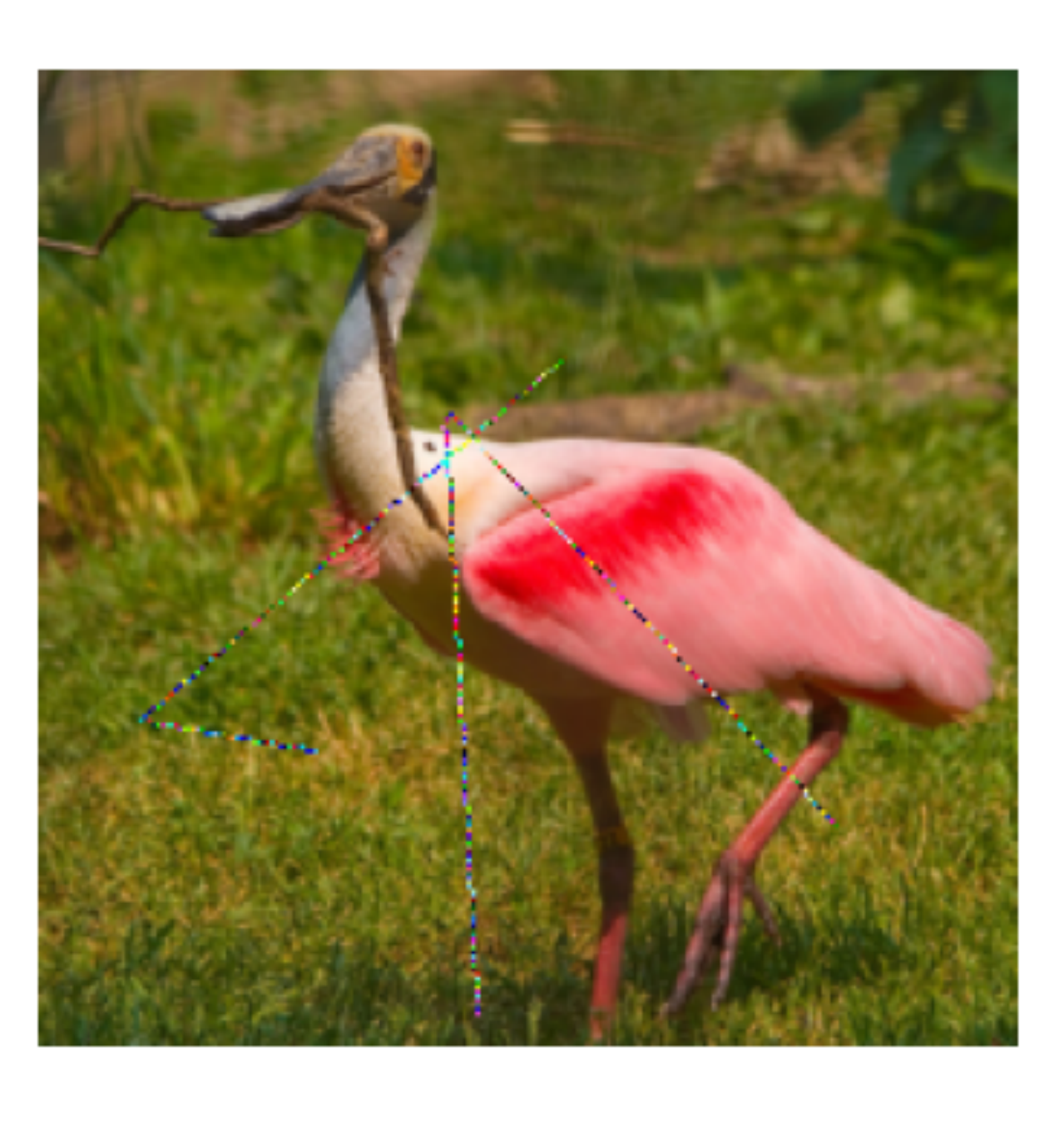}
    \end{subfigure}
    \\[5pt]
    \begin{subfigure}{.5\linewidth}
      \centering
    %   \footnotesize \textsf{Original Image \\ porcupine, hedgehog \\[-5pt]}
      \caption*{Porcupine, hedgehog}
      \vspace{-1.2em}
      \includegraphics[clip, trim=4cm 8cm 4cm 8cm, width=0.85\linewidth]{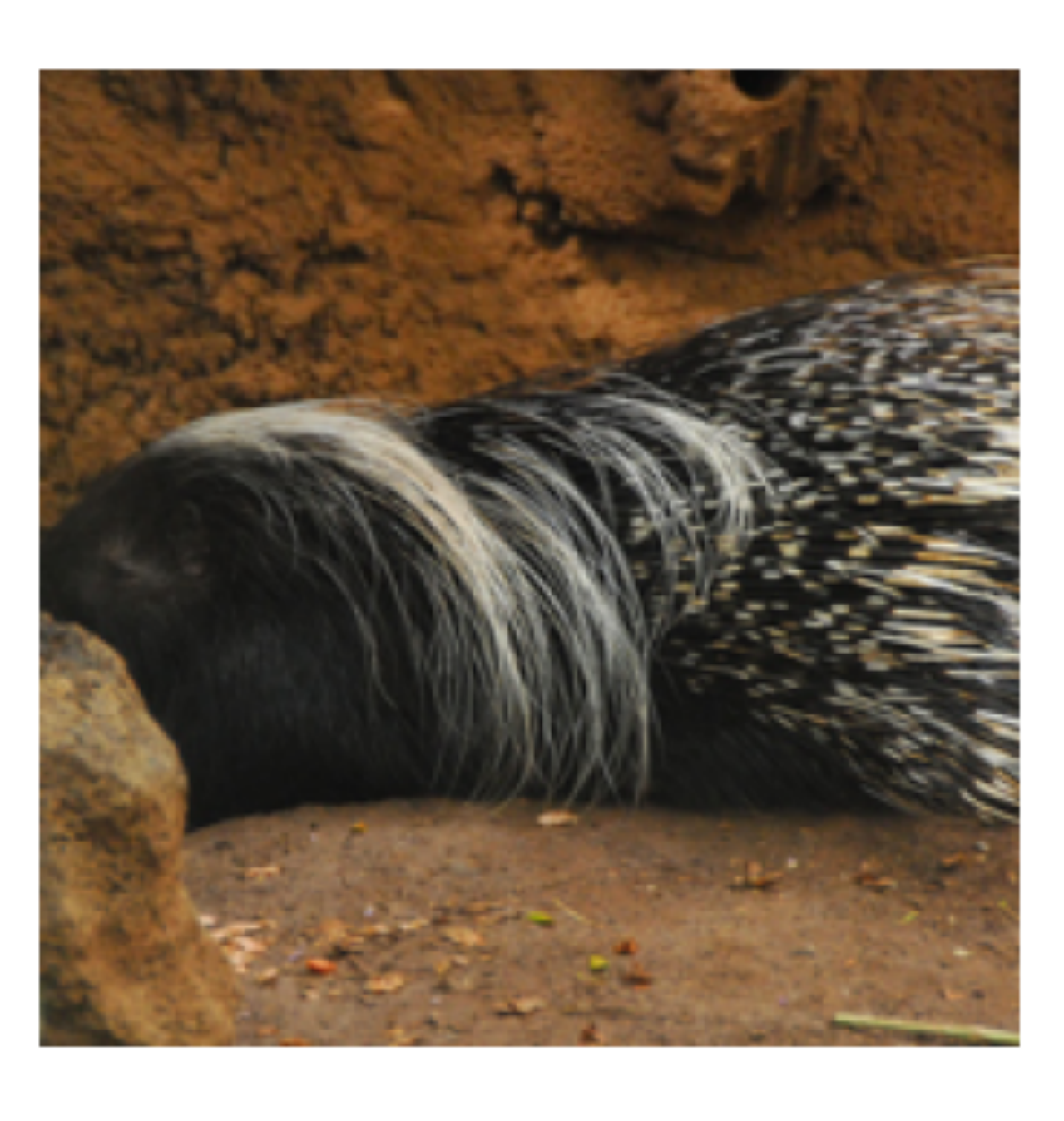}
    \end{subfigure}%
    \hspace{-2em}
    \begin{subfigure}{.5\linewidth}
      \centering
    %   \footnotesize \textsf{Adversarial Image \\ church, church building \\[-5pt]}
      \caption*{Church, church building}
      \vspace{-1.2em}
      \includegraphics[clip, trim=4cm 8cm 4cm 8cm, width=0.85\linewidth]{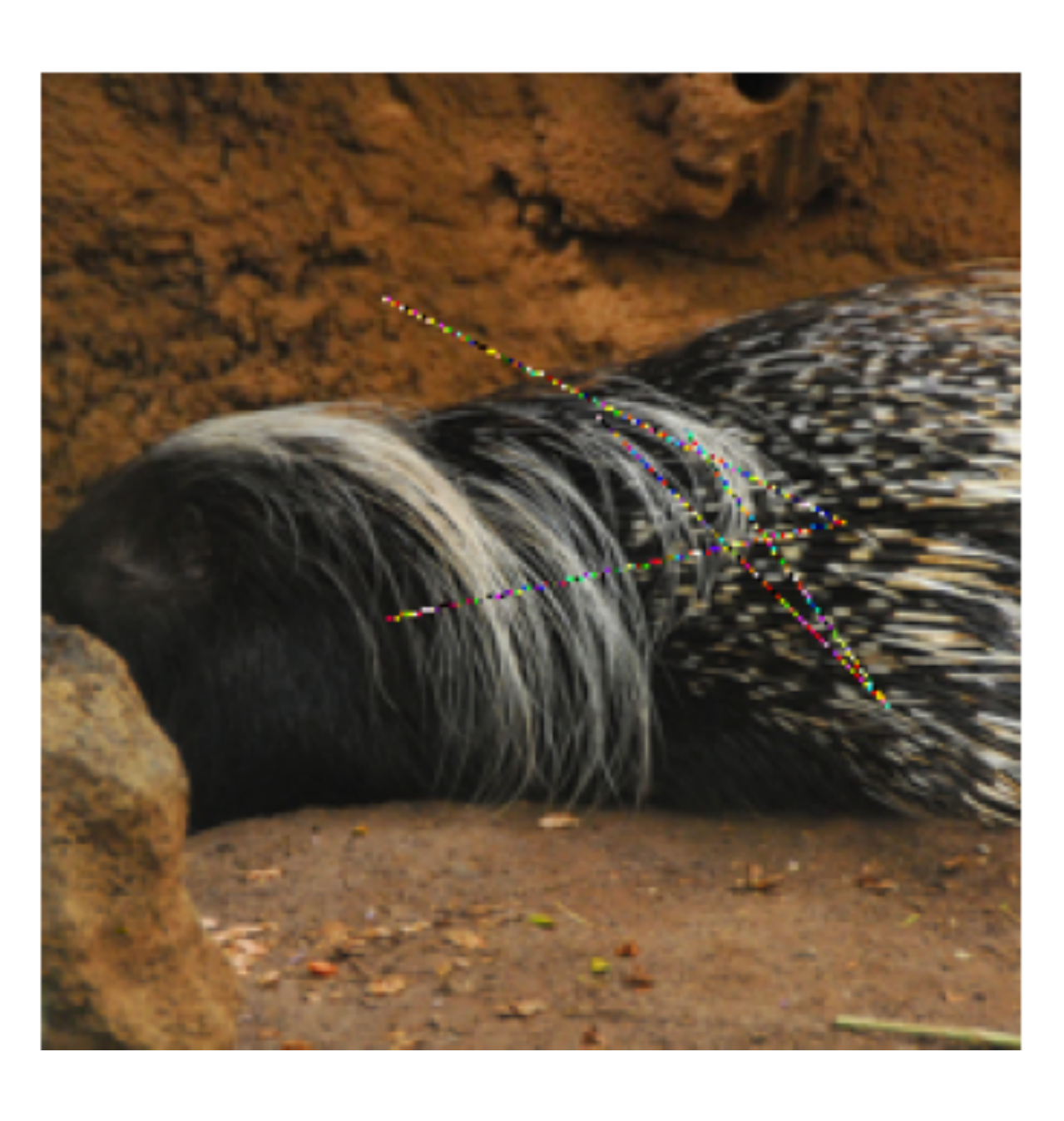}
    \end{subfigure}
    
    \caption{Examples of CMA-ES B\a'ezier curves attacks in the network domain. Scratches were clipped to the image domain for visualization purposes. The left side shows the original images and their corresponding class; the right side shows the generated adversarial samples (scratches) and their respective predicted labels when evaluated on ResNet-50. Best viewed in color.}
    \label{fig:network-bezier}
\end{figure}
\noindent\textbf{B\a'ezier curves attacks.}
We changed the shape of the adversarial scratches, this times using quadratic B\a'ezier curves, as explained in Section \ref{sec:image-domain-attack}. Results are reported in Tables~\ref{tab:network_domain_fixed_scratch_imagenet} and \ref{tab:network_domain_fixed_scratch_cifar10}, respectively. We can see again that increasing the number of scratches decreases the number of queries of a successful attack and increases its success rate. Moreover, we can observe that B\a'ezier curves attacks are more effective than the corresponding straight lines attacks, in terms of the number of queries needed and the success rate. Our intuition is that the attacked model uses more information on the shape of the image rather than its colors when making a prediction; hence, a nonlinear scratch like a B\a'ezier curve can better exploit this behavior.
This may also explain why a higher number of scratches results in a higher-success rate: scratches located in different parts of an image can exploit more shape-dependent features.

\begin{table*}[h]
\centering
\begin{tabular}{cccccccc}
\hline
\multicolumn{8}{c}{\textbf{Network Domain}} \\ \hline
\multicolumn{4}{c|}{\textbf{Bézier Curves}} & \multicolumn{4}{c}{\textbf{Straight Lines}} \\ \hline
\textbf{Scratches} & \textbf{Success Rate} & \textbf{Queries} & \multicolumn{1}{c|}{\textbf{Coverage}} & \textbf{Scratches} & \textbf{Success Rate} & \textbf{Queries} & \textbf{Coverage} \\ \hline
\multicolumn{1}{c|}{1} & \multicolumn{1}{c|}{53.80\%} & \multicolumn{1}{c|}{31718} & \multicolumn{1}{c|}{0.46\%} & \multicolumn{1}{c|}{1} & \multicolumn{1}{c|}{44.00\%} & \multicolumn{1}{c|}{34623} & 0.56\% \\
\multicolumn{1}{c|}{2} & \multicolumn{1}{c|}{86.40\%} & \multicolumn{1}{c|}{25747} & \multicolumn{1}{c|}{0.84\%} & \multicolumn{1}{c|}{2} & \multicolumn{1}{c|}{80.50\%} & \multicolumn{1}{c|}{30776} & 0.99\% \\
\multicolumn{1}{c|}{3} & \multicolumn{1}{c|}{96.70\%} & \multicolumn{1}{c|}{16838} & \multicolumn{1}{c|}{1.24\%} & \multicolumn{1}{c|}{3} & \multicolumn{1}{c|}{92.40\%} & \multicolumn{1}{c|}{23826} & 1.42\% \\
\multicolumn{1}{c|}{4} & \multicolumn{1}{c|}{98.77\%} & \multicolumn{1}{c|}{11640} & \multicolumn{1}{c|}{1.65\%} & \multicolumn{1}{c|}{4} & \multicolumn{1}{c|}{96.40\%} & \multicolumn{1}{c|}{18848} & 1.86\% \\ \hline
\end{tabular}
\caption{Network domain, fixed-location scratch attack success rate for ResNet-50 trained on ImageNet.}
\label{tab:network_domain_fixed_scratch_imagenet}
\end{table*}

\begin{table*}[h]
\centering
\begin{tabular}{cccccccc}
\hline
\multicolumn{8}{c}{\textbf{Network Domain}} \\ \hline
\multicolumn{4}{c|}{\textbf{Bézier Curves}} & \multicolumn{4}{c}{\textbf{Straight Lines}} \\ \hline
\textbf{Scratches} & \textbf{Success Rate} & \textbf{Queries} & \multicolumn{1}{c|}{\textbf{Coverage}} & \textbf{Scratches} & \textbf{Success Rate} & \textbf{Queries} & \textbf{Coverage} \\ \hline
\multicolumn{1}{c|}{1} & \multicolumn{1}{c|}{77.80\%} & \multicolumn{1}{c|}{2697} & \multicolumn{1}{c|}{2.97\%} & \multicolumn{1}{c|}{1} & \multicolumn{1}{c|}{54.90\%} & \multicolumn{1}{c|}{3133} & 3.28\% \\
\multicolumn{1}{c|}{2} & \multicolumn{1}{c|}{93.10\%} & \multicolumn{1}{c|}{2089} & \multicolumn{1}{c|}{4.21\%} & \multicolumn{1}{c|}{2} & \multicolumn{1}{c|}{76.80\%} & \multicolumn{1}{c|}{3207} & 3.96\% \\
\multicolumn{1}{c|}{3} & \multicolumn{1}{c|}{96.00\%} & \multicolumn{1}{c|}{2000} & \multicolumn{1}{c|}{4.54\%} & \multicolumn{1}{c|}{3} & \multicolumn{1}{c|}{83.80\%} & \multicolumn{1}{c|}{2829} & 4.30\% \\
\multicolumn{1}{c|}{4} & \multicolumn{1}{c|}{97.20\%} & \multicolumn{1}{c|}{1915} & \multicolumn{1}{c|}{4.69\%} & \multicolumn{1}{c|}{4} & \multicolumn{1}{c|}{90.70\%} & \multicolumn{1}{c|}{2622} & 4.40\% \\ \hline
\end{tabular}
\caption{Network domain, fixed-location scratch attack success rate for ResNet-50 trained on CIFAR-10.}
\label{tab:network_domain_fixed_scratch_cifar10}
\end{table*}

\subsection{Variable Location Network Domain}
Network-domain attacks with variable locations achieve lower success rates than their image-domain counterparts, as illustrated in Table~\ref{tab:network_domain_variable_scratch_cifar10}. We hypothesize that this may be due to searching over a much larger search space when taking into account the relaxation on the pixel value restrictions. Furthermore, we observe that the majority of variable-location attacks on ImageNet time out with our query budget, motivating the need for a much larger query budget for such attacks.

\begin{table*}[h]
\centering
\begin{tabular}{cccccccc}
\hline
\multicolumn{8}{c}{\textbf{Network Domain}} \\ \hline
\multicolumn{4}{c|}{\textbf{Bézier Curves}} & \multicolumn{4}{c}{\textbf{Straight Lines}} \\ \hline
\textbf{Scratches} & \textbf{Success Rate} & \textbf{Queries} & \multicolumn{1}{c|}{\textbf{Coverage}} & \textbf{Scratches} & \textbf{Success Rate} & \textbf{Queries} & \textbf{Coverage} \\ \hline
\multicolumn{1}{c|}{1} & \multicolumn{1}{c|}{52.18\%} & \multicolumn{1}{c|}{1578} & \multicolumn{1}{c|}{4.15\%} & \multicolumn{1}{c|}{1} & \multicolumn{1}{c|}{63.49\%} & \multicolumn{1}{c|}{350} & 1.98\% \\
\multicolumn{1}{c|}{2} & \multicolumn{1}{c|}{68.25\%} & \multicolumn{1}{c|}{886} & \multicolumn{1}{c|}{6.57\%} & \multicolumn{1}{c|}{2} & \multicolumn{1}{c|}{68.45\%} & \multicolumn{1}{c|}{410} & 3.94\% \\
\multicolumn{1}{c|}{3} & \multicolumn{1}{c|}{75.77\%} & \multicolumn{1}{c|}{940} & \multicolumn{1}{c|}{8.62\%} & \multicolumn{1}{c|}{3} & \multicolumn{1}{c|}{72.22\%} & \multicolumn{1}{c|}{1340} & 5.1\% \\ \hline
\end{tabular}
\caption{Network domain, variable-location scratch attack success rate for ResNet-50 trained on CIFAR-10. We use differential evolution with relaxed constraints on the pixel values.}
\label{tab:network_domain_variable_scratch_cifar10}
\end{table*}

% \begin{table}[h]
%     \centering
%     \begin{tabular}{| c | C{2cm} | c | c |}
%         \hline
%         Scratches & Mean pixels covered & Queries & Success rate \\ \hline
%         1 & 2.97\% & 2697 & 77.80\% \\ \hline
%         2 & 4.21\% & 2089 & 93.10\% \\ \hline
%         3 & 4.54\% & 2000 & 96.00\% \\ \hline
%         4 & 4.69\% & 1915 & 97.20\% \\ \hline
%     \end{tabular}
%     \caption{CIFAR10, Bézier curves attacks}
%     \label{table:cifar10-bezier}
% \end{table}

% \begin{table}[h]
%     \centering
%     \begin{tabular}{| c | C{2cm} | c | c |}
%         \hline
%         Scratches & Mean pixels covered & Queries & Success rate \\ \hline
%         1 & 0.46\% & 31718 & 53.80\% \\ \hline
%         2 & 0.84\% & 25747 & 86.40\% \\ \hline
%         3 & 1.24\% & 16838 & 96.70\% \\ \hline
%         4 & 1.65\% & 11640 & 98.77\% \\ \hline
%     \end{tabular}
%     \caption{ImageNet, Bézier curves attacks}
%     \label{table:imagenet-bezier}
% \end{table}

\section{Discussion}
\label{sec:discussion}

\subsection{Unsuccessful Image-Domain Scratch Attempts} Prior to obtaining our results, we experimented with a broad range of parameters for the scratches. For the image domain, we attempted the following:
\begin{itemize}
    \item \textbf{CMA-ES.} We initially used CMA-ES with fixed scratch locations and \textit{tanh} normalization to implement image-domain attacks similar to the CW attack~\cite{carlini2017towards}. However, this yielded an attack success of around $2\%$, which motivated us to use Differential Evolution, which significantly improved attack success rate. 
    \item \textbf{Fixed scratch location.} We initially attempted our image-domain attacks with fixed scratch parameters and evolved only the RGB values. This led to our fitness functions often getting stuck in local minima and not converging to the target class; hence, we included the scratch location parameters in the evolution strategy. 
\end{itemize}

\subsection{Source-target Dependency in Network Domain}

\begin{figure}[h!]
    \centering
    \includegraphics[clip, trim=0.3cm 2cm 0.3cm 1.5cm, width=\linewidth]{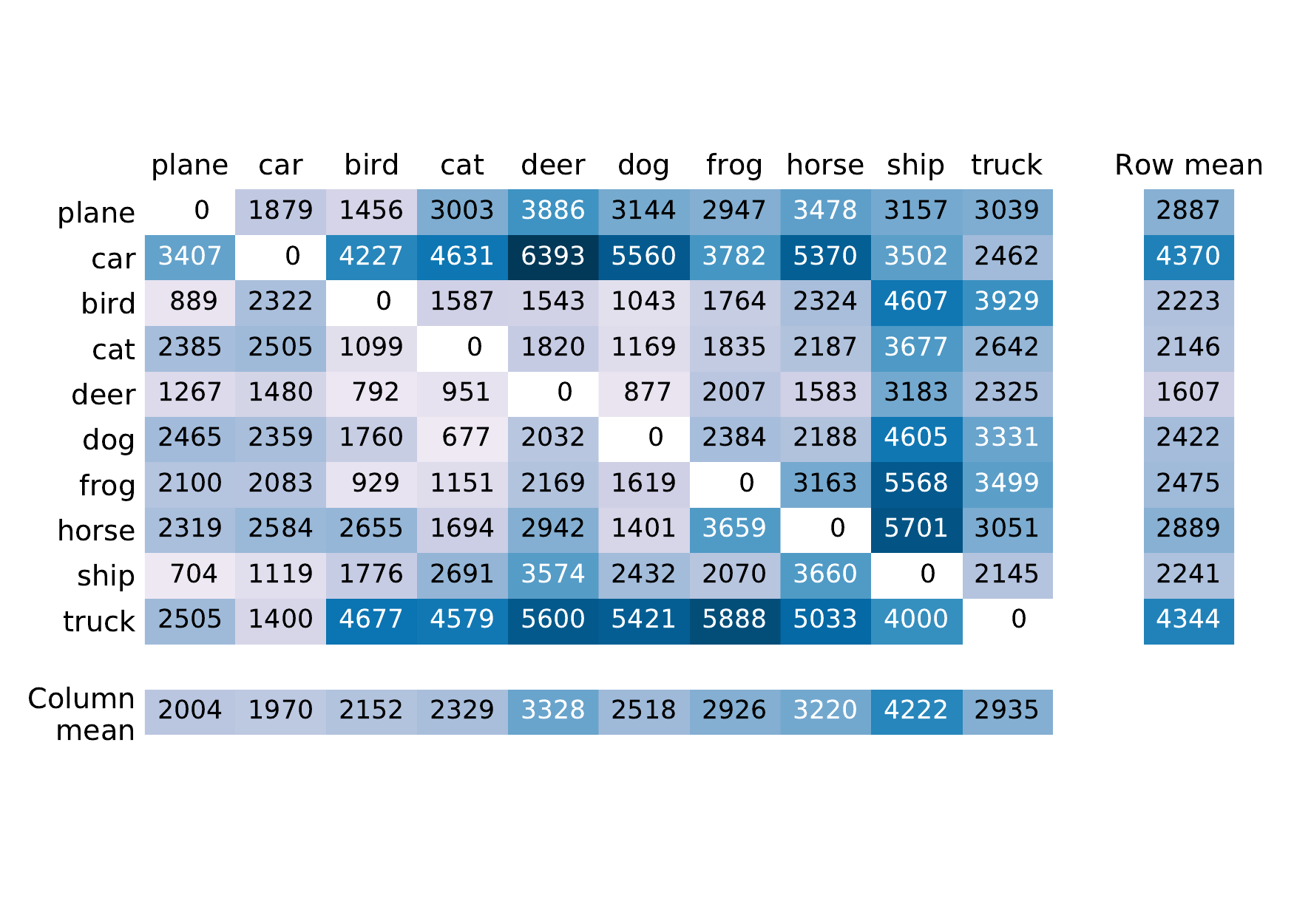}
    \caption{Source-target dependency on CIFAR-10. Cell values correspond to the average number of queries to attack a source class (row) with a target class (column). The figure also shows the means of each row and column.}
    \label{fig:src-target}
\end{figure}

We wanted to analyze whether there was a relationship between the ground truth and the target class with respect to the difficulty of the attack. 
% We wanted to analyze whether the choice of the target class, given a source image, can have an influence on the difficulty of the attack. 
For practical reasons, we performed the following experiments on CIFAR-10, given that it has only 10 classes compared to ImageNet with 1000 classes.
The results in Figure~\ref{fig:src-target} are based on 500 attacked images with 50 samples per target class.
% Attack settings: Network domain, 4 scratches, 5% of maximum coverage
In this figure, rows correspond to source classes, columns correspond to target classes, and cell values correspond to the average number of queries needed to achieve a successful attack.

The mean value for each row, displayed in the right-most column in Figure~\ref{fig:src-target}, provides the average number of queries needed to perform a successful attack for images with that particular source class.
% , and the right-most column in Figure~\ref{fig:src-target} highlights the mean queries for that source class. 
We observe that not all source classes are equally easy to attack; the \texttt{deer} class is the easiest to attack, while the \texttt{car} and \texttt{truck} classes are the hardest. 
% This means that given a dataset and a type of attack, some classes are easier to attack than others.
%
Conversely, the mean values for each column, displayed in the bottom row of Figure~\ref{fig:src-target}, provide the average number of queries required to generate a successful adversarial sample with that class as the target. We observe that not all classes are easy to reach as targets; the \texttt{ship} and \texttt{deer} classes require the most queries when chosen as targets.
% On the other hand, the mean value for each column provides the average number of queries needed by each target class to perform a successful attack. The bottom-most row in Figure~\ref{fig:src-target} shows the results. 
%
% \highlight{Even in this scenario, some of the target classes perform better than others regarding the number of queries needed to attack any of the source classes.}
% \bh{It would be nice to move Fig.~\ref{fig:src-target} close to this paragraph. It feels like one might lose track of what `source' and `target' class mean.}
%
% As expected, some target classes perform better than others in terms of number of queries to attack any source class of a given dataset. In our results, the \texttt{car} and \texttt{plane} classes are the most effective, while the \texttt{ship} class is the worst. Hence, the choice of the target class has an impact on the effectiveness of the attack.
%
% \lr{The technical reviewer was confused about the concept of 'effectiveness' here above. I tried to add more clarifying information hoping our future reviewers will get it correctly. Please see if it needs further improvements}

\begin{figure}[h]
    \centering
    \includegraphics[clip, trim=0.3cm 2cm 0.3cm 1cm, width=\linewidth]{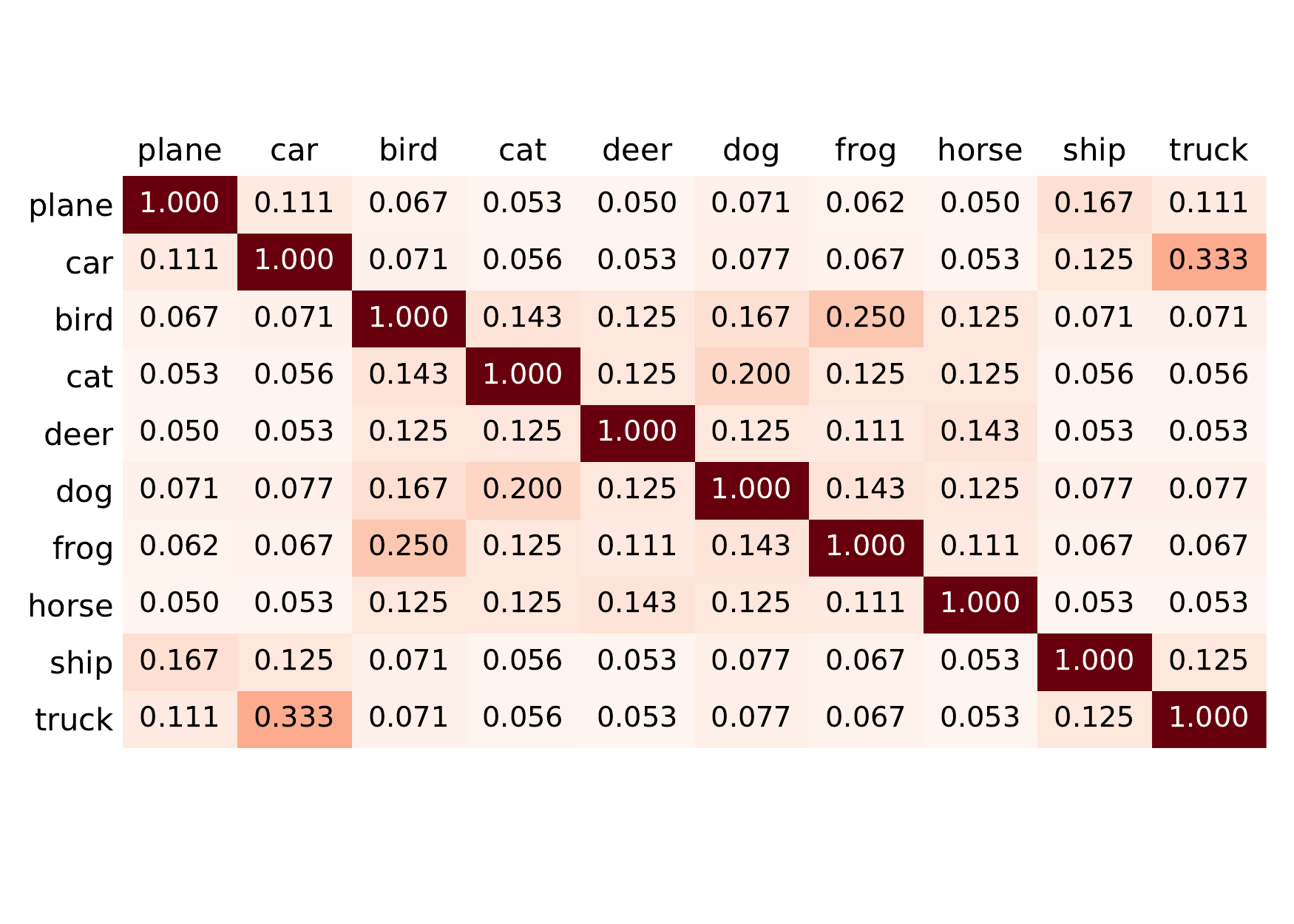}
    \caption{Similarity of CIFAR-10 classes computed with WordNet.}
    \label{fig:similarities}
\end{figure}

Furthermore, we attempted to investigate the semantic similarity between source and target classes. We computed word similarity measures using the WordNet dataset~\cite{wordnet}, with the results highlighted in Figure~\ref{fig:similarities}. Given the results in Figure~\ref{fig:src-target}, we conjecture that
higher similarity between the source and target labels typically results in a more effective attack. For example, for the (\texttt{truck}, \texttt{car}) pair, we need relatively fewer queries to generate targeted attacks towards \texttt{car} classes from a \texttt{truck} image, and vice versa. This phenomena persists for other pairs with similar semantic information, such as (\texttt{cat}, \texttt{dog}) or (\texttt{bird},\texttt{plane}).
% attacking a \texttt{truck} image with the target \texttt{car} \lr{The technical reviewer didn't get the meaning of 'attacking a truck image with the target car'. Any ideas on how to re-phrase this?} is the best choice, and the vice-versa holds as well. 

However, we believe that this query dependency is tightly connected with the method used to compute class similarity; for instance, while \texttt{bird} and \texttt{plane} classes share similarity in that they represent flying objects, that is not the case for pairs such as (\texttt{truck}, \texttt{car}) and  (\texttt{dog}, \texttt{cat}), as illustrated in Figure~\ref{fig:similarities}. One possible conjecture might be that although semantic similarity between classes plays a major role in the ease of an attack, simple tree-based structures such as WordNet are not sufficient. Further work is needed to devise suitable semantic similarity metrics between source and target classes.

 %this behavior may not be manifested in WordNet,

% (\texttt{bird}, \texttt{plane}).
% Therefore, we conjecture that attacking an image using a target class similar to the image class is more effective than using a target that is very different.

\subsection{Defenses in Image Domain}
\label{sec:defenses-img-domain}

\noindent\textbf{Image Filtering.}
We discuss several possible image filter solutions to prevent image domain adversarial-scratch attacks. We evaluated the following techniques as defenses against adversarial scratched images in the image domain:
\begin{itemize}
    \item JPEG compression with varying quality factors
    \item Median filters applied to each channel separately with a kernel size of $3\times3$ pixels
\end{itemize}

\begin{figure}[h!]
    \centering
    \begin{subfigure}{.5\linewidth}
      \centering
      {\footnotesize Original Image}
      \includegraphics[width=0.85\linewidth]{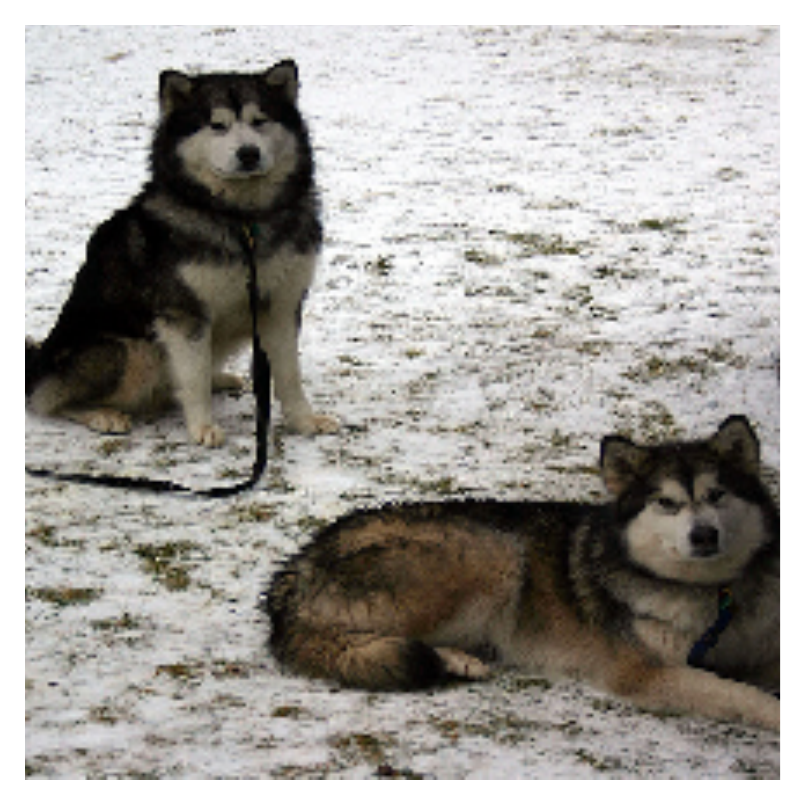}
    \end{subfigure}%
    \hspace{-2em}
    \begin{subfigure}{.5\linewidth}
      \centering
      {\footnotesize Adversarial Image}
      \includegraphics[width=0.85\linewidth]{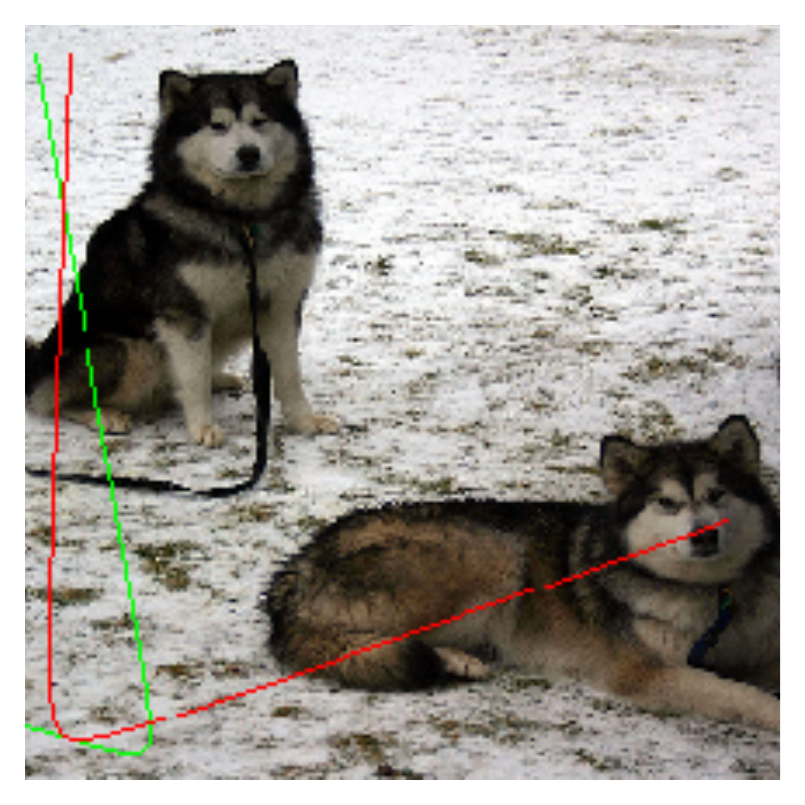}
    \end{subfigure}
    \\[5pt]
    
    \begin{subfigure}{.5\linewidth}
      \centering
      {\footnotesize Adversarial Image \\ JPEG compressed}
      \includegraphics[width=0.85\linewidth]{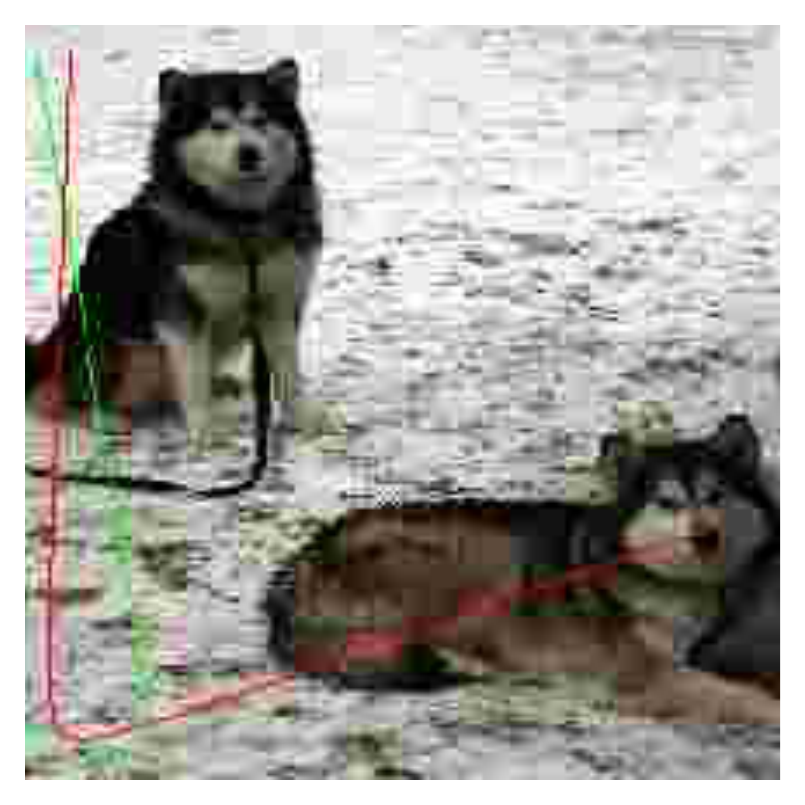}
    \end{subfigure}%
    \hspace{-2em}
    \begin{subfigure}{.5\linewidth}
      \centering
      {\footnotesize Adversarial Image \\ Median Filtered}
      \includegraphics[width=0.85\linewidth]{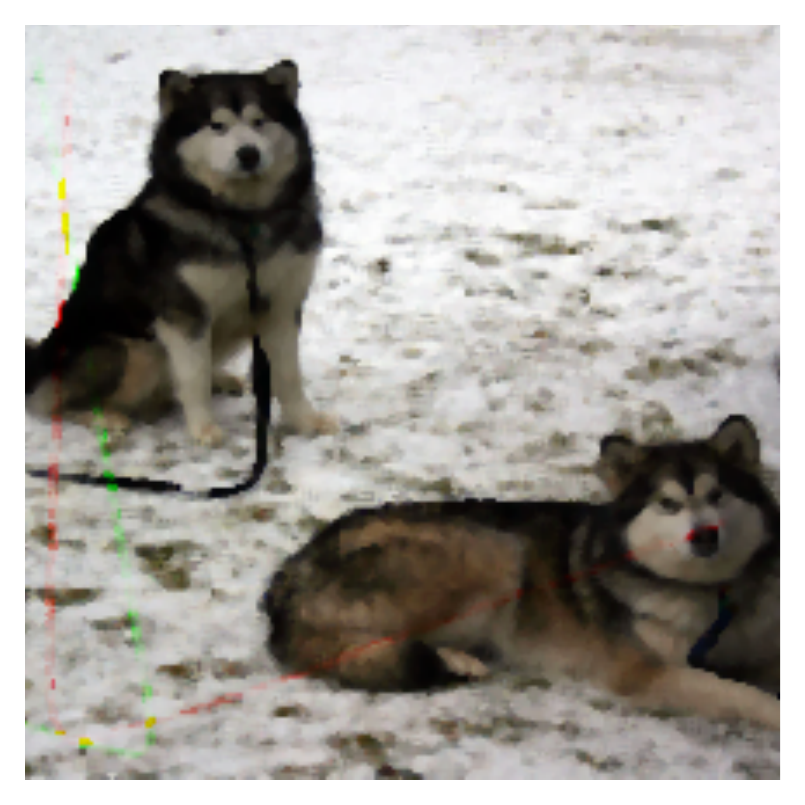}
    \end{subfigure}

    \caption{Best viewed in color. Filtering methods to counter adversarial noise. We visualize the results of various image compression mechanisms, namely JPEG compression and median filtering. JPEG compression fails to remove the scratches in an image, which might be responsible for its ineffectiveness as a defense. Median filtering on the other hand allows for much better smoothing and removal of the scratches.}
    % \bh{This figure is confusing. Could you elaborate a bit more on it, e.g., showing that the target class does not get affected after adding the defense or sth. similar?}}
    \label{fig:img_domain_compression}
\end{figure}

We capture this behavior through the recovery rate, defined as the fraction of adversarial images that are predicted according to their correct labels after an adversarial scratches attack.
Surprisingly, we find that JPEG compression is highly ineffective as a defense against image-domain scratches targeting ImageNet-trained ResNet-50 models. We observe that we recovered $0\%$ of original labels from adversarially scratched images. Correspondingly, JPEG compression also leads to a low recovery rate for CIFAR-10 trained ResNet-50 models. Median filtering offers a better recovery rate, as highlighted in Table~\ref{tab:img_domain_resnet50_defenses}. We observe visually in Figure~\ref{fig:img_domain_compression} that median filtering can physically remove scratch pixels from the image, while JPEG compression often fails to do so, which might explain median filtering's effectiveness as a defense.

\begin{table}[h!]
\centering
\begin{tabular}{|c|C{2.3cm}|C{2.3cm}|}
\hline
\textbf{Method} & \textbf{ImageNet Recovery Rate} & \textbf{CIFAR-10 Recovery Rate} \\ \hline
JPEG, q=85 & 0\% & 30.4\% \\ \hline
JPEG, q=90 & 0\% & 36.58\% \\ \hline
JPEG, q=95 & 0\% & 34.14\% \\ \hline
JPEG, q=99 & 0\% & 24.3\% \\ \hline
Median filter & 100\% & 47.56\% \\ \hline
\end{tabular}
\caption{Recovery rate for ImageNet-trained ResNet-50 against double scratch Bézier curve attacks.}
\label{tab:img_domain_resnet50_defenses}
\end{table}

\noindent\textbf{Certified Defenses.}
Furthermore, we evaluated the image-domain attack against certified defenses for adversarial patches~\cite{chiang2020certified}, which provides a certificate of robustness, or an upper bound on the attack success rate for a given test set and $L_{0}$-based threat model, which is most suited for our attack. We did not evaluate our attack against $L_{\infty}$ or $L_{2}$ certified defenses as they lie outside our threat model. We trained a Convolutional Neural Network (CNN) from Chiang et al.~\cite{chiang2020certified} with four
convolutional layers with kernel size (3, 4, 3, 4), stride (1, 2, 1, 2), output channels (4, 4, 8 ,8), and two
fully connected layers with $256$ neurons on CIFAR-10. It achieved a test accuracy of $36.9\%$ and a certified accuracy of $21.90\%$ for $L_{0}$ attacks that lie within a $5\times5$ pixel patch. 

Our attack on this network achieves a $2.2\%$ success rate for $1000$ CIFAR-10 images when taking into account the threat model of adversarial scratches falling within a $5\times5$ patch, and $34\%$ when we remove this restriction. We find certified defenses to be a suitable defense against our attack. However, they are computationally intensive and  currently cannot be scaled to larger neural networks such as those trained on ImageNet. Training certifiably robust neural networks for larger architectures remains an open problem.

\subsection{Defenses in Network Domain}
\label{sec:defenses-network-domain}
Here we discuss possible solutions to prevent the proposed attacks in the network domain.
We evaluated the following techniques as defenses against adversarial scratched images:
\begin{itemize}
    \item Clipping to the range [0, 1]; i.e., clip each pixel of the given image so that they fall into the image domain
    \item JPEG compression with varying quality factors
    \item Median filter applied to each channel separately with a kernel size of 3x3 pixels
\end{itemize}

To assess their performance, we use the notion of recovery rate defined in Section \ref{sec:defenses-img-domain}. Interestingly, we found out that about 21\% of the attacks on CIFAR-10, even if we are in the network domain, fall in the image domain. 
% At the end of this section we provide an explanation for this phenomenon. 
All attacks on ImageNet fall outside the image domain. Tables~\ref{table:cifar10-defenses} and \ref{table:imagenet-defenses} show the recovery rate with respect to all the available adversarial attacks. Table~\ref{table:cifar10-defenses} also shows the recovery rate after splitting the attacks into the ones that are in the network domain and the ones that fall in the image domain to evaluate the performance of the defenses on each domain separately.
%\gc{This is not clear just like the caption - relative to the domain itself?}

\begin{table}[h]
    \centering
    \begin{tabular}{| l | C{1.55cm} | C{1.3cm} | C{1.2cm} |}
        \hline
        \multicolumn{1}{|c|}{\textbf{Method}} & \textbf{Recovery Rate} & \textbf{Network Domain} & \textbf{Image Domain} \\ \hline
        Clipping                & 53\% & 70\% & 0\% \\ \hline
        JPEG, quality = 90      & 77\% & 86\% & 47\% \\ \hline
        JPEG, quality = 95      & 80\% & 90\% & 49\% \\ \hline
        JPEG, quality = 99      & 82\% & 91\% & 51\% \\ \hline
        Median filter           & 86\% & 88\% & 81\% \\ \hline
    \end{tabular}
    \caption{Performance of defense methods applied to network-domain fixed-location attacks on CIFAR-10. We highlight the recovery rates after performing image filtering on image domain scratches obtained from network domain attacks.} 
    % \lr{Not sure this is well written, feel free to improve it.} }
    \label{table:cifar10-defenses}
\end{table}

\begin{table}[h]
    \centering
    \begin{tabular}{| l | c | c | c |}
        \hline
        \multicolumn{1}{|c|}{\textbf{Method}} & \textbf{Recovery Rate} \\ \hline
        Clipping                & 66\% \\ \hline
        JPEG, quality = 85      & 81\% \\ \hline
        JPEG, quality = 90      & 79\% \\ \hline
        JPEG, quality = 95      & 76\% \\ \hline
        JPEG, quality = 99      & 76\% \\ \hline
        Median filter           & 78\% \\ \hline
    \end{tabular}
    \caption{Performance of defense methods applied to network-domain fixed-location attacks on ImageNet.}
    \label{table:imagenet-defenses}
\end{table}

%We also evaluated the performance impact when we apply the defense methods over the testset. Results are shown in Tables~\ref{table:cifar10-drop} and \ref{table:imagenet-drop}.
Tables~\ref{table:cifar10-drop} and \ref{table:imagenet-drop} show the performance impacts after applying the defense methods over the testset.
\begin{table}[h]
    \centering
    \begin{tabular}{| l | c | c | c |}
        \hline
        \multicolumn{1}{|c|}{\textbf{Method}} & \textbf{Accuracy} & \textbf{Drop} \\ \hline
        No defense method  & 93.91\% & --- \\ \hline
        Clipping           & 93.91\% & 0.00\% \\ \hline
        JPEG, quality = 99 & 93.72\% & 0.19\% \\ \hline
        Median filter      & 80.00\% & 13.91\% \\ \hline
    \end{tabular}
    \caption{Drop of ResNet-50 model performance after applying defense methods on CIFAR-10 testset.}
    \label{table:cifar10-drop}
\end{table}
\begin{table}[h]
    \centering
    \begin{tabular}{| l | c | c | c |}
        \hline
        \multicolumn{1}{|c|}{\textbf{Method}} & \textbf{Accuracy} & \textbf{Drop} \\ \hline
        No defense method  & 76.13\% & --- \\ \hline
        Clipping           & 76.13\% & 0.00\% \\ \hline
        JPEG, quality = 85 & 74.55\% & 1.58\% \\ \hline
        Median filter      & 71.97\% & 4.16\% \\ \hline
    \end{tabular}
    \caption{Drop of ResNet-50 model performance after applying defense methods on ImageNet testset.}
    \label{table:imagenet-drop}
\end{table}

Considering both the effectiveness of the defense and the drop in model performance on benign images, JPEG compression appears to be the best defense in the network domain, due to its low impact on model performance and highest prevention rate, while median filtering appears to be the best defense for image domain attacks. 
% We can see in Table~\ref{table:cifar10-defenses} that the median filter has a much higher prevention rate in the image domain with respect to other techniques, confirming the good performance obtained in Section~\ref{sec:defenses-img-domain}.

% \begin{figure}[h!]
%     \centering
%     \includegraphics[clip, trim=2.9cm 3cm 2.8cm 3cm, width=.85\linewidth]{./figures/experiments/net-img-domain.pdf}
%     \caption{Highlight of network-domain, fixed-location attacks on CIFAR-10 that actually fall within the image domain.}
%     \label{fig:net-img-domain}
% \end{figure}

%\vspace{0.1in}
% \noindent\textbf{Why Some Attacks Fall in Image Domain.}
%\label{sec:why-image-domain}\gc{label will not work}
% Some of the network-domain attacks on CIFAR-10 are actually in the image domain. In Figure~\ref{fig:net-img-domain}, we observe that image domain attacks need much fewer queries with respect to the attacks in the network domain. This is reasonable; the image-domain attacks did not go far from the initialized values of the CMA-ES algorithm. 
We can appreciate that the majority of image-domain attacks need fewer than 400 queries, corresponding to at most 10 iterations of the CMA-ES algorithm with a population size of 40, as described in Section~\ref{sec:network-domain-setup}.

\section{Conclusion}
\label{sec:conclusion}

% In this paper, we proposed a new attack against neural networks, called \textit{adversarial scratches}. 
In this paper, we proposed \emph{adversarial scratches}, a new attack against neural networks inspired by the scratches that are commonly found in old printed photographs.

Our attack operates under strict assumptions: adversaries have access only to the predictions of a model (i.e, black-box access) and have limited ability in altering the input data. Our attack proved to be effective against not only several state-of-the-art deep neural network image classifiers, but also Microsoft's Cognitive Services Image Captioning API. To offer a complete perspective, we also proposed mitigation strategies against our attack and demonstrated that we have the ability to restore the performance of the attacked models.

As future work, we aim to gain a better understanding of the correlation between source-target classes, their potential semantic similarities, and the success of the adversarial scratches attack. Furthermore, building upon the preliminary results obtained in this paper, we intend to expand our investigation on potential mitigation strategies. Additionally, we will expand our attack beyond the image classification. For instance, the equivalent of adversarial
scratches for speech recognition systems can be posed as chirped
sequences that appear benign, but might potentially be malicious.

\bibliographystyle{plain}
\balance
\bibliography{bibliography}

%%%%%%%%%%%%%%%%%%%%%%%%%%%%%%%%%%%%%%%%%%%%%%%%%%%%%%%%%%%%%%%%%%%%%%%%%%%%%%%%
\end{document}